\documentclass[final]{cvpr}

\usepackage{times}
\usepackage{epsfig}
\usepackage{microtype}
\usepackage{graphicx}
\usepackage[font=small]{caption}
\usepackage[font=small]{subcaption}
\usepackage{xspace}
\usepackage{amsmath, bm}
\usepackage{amssymb}
\usepackage{algorithm}
\usepackage{algorithmic}
\usepackage{mathtools}
\usepackage{array}
\usepackage{multirow}
\usepackage[inline]{enumitem}
\usepackage{multirow}
\usepackage{tabularx}
\usepackage{color}
\usepackage[utf8]{inputenc} 
\usepackage[T1]{fontenc}    
\usepackage[hyphens]{url}   
\usepackage{booktabs}       
\usepackage{amsfonts}       
\usepackage{xcolor}
\usepackage{blindtext}
\usepackage[numbers,sort]{natbib}
\usepackage{flushend}
\usepackage{booktabs}
\usepackage{lipsum}
\usepackage{bbm}

\usepackage[pagebackref=true,breaklinks=true,colorlinks,bookmarks=false]{hyperref}

\begin{document}

\title{Zero-Shot Learning with Knowledge Enhanced Visual Semantic Embeddings}

\author{
Karan Sikka$^1$\thanks{Corresponding author is Karan Sikka (karan.sikka@sri.com).} \;\; Jihua Huang$^1$ \;\;  Andrew Silberfarb$^1$ \;\;  Prateeth Nayak$^1$ \;\;  Luke Rohrer$^{1, 2}$\thanks{This work was done as an intern at SRI International.}\\
Pritish Sahu$^{3}$\footnotemark[2] \;\; John Byrnes$^1$ \;\;  Ajay Divakaran$^1$ \;\;  Richard Rohwer$^1$ \\
$^1$SRI International\\
$^2$University of California San Diego\\
$^3$Rutgers University\\
}


\maketitle

\def\@onedot{\ifx\@let@token.\else.\null\fi\xspace}
\makeatother
\def\etal{\emph{et al}\onedot}
\def\etc{\emph{etc}\onedot}
\def\ie{\emph{i.e}\onedot}
\def\eg{\emph{e.g}\onedot}
\def\cf{\emph{cf}\onedot}
\def\vs{\emph{vs}\onedot}
\def\pd{\partial}
\def\grad{\nabla}
\def\R{\mathbb{R}}
\def\G{\mathbb{G}}
\def\d{\boldsymbol{\delta}}
\def\y{\textbf{y}}
\def\l{\boldsymbol{\ell}}
\def\wrt{w.r.t\onedot}
\def\a{\boldsymbol{\alpha}}
\def\vertspace{0.6em}
\newcommand{\mat}[1]{\bm{#1}}
\newcommand{\set}[1]{\mathbb{#1}}
\newcommand{\colons}[1]{``{#1}''}
\def\pytorch{PyTorch}
\newcommand{\norm}[1]{||{#1}||_2}

\def\mF{\mat{F}}
\def\mX{\mat{X}}
\def\mS{\mat{S}}
\def\mM{\mat{M}}
\def\mW{\mat{W}}
\def\vs{\mat{s}}
\def\vx{\mat{x}}
\def\ve{\mat{e}}
\def\vy{\mat{y}}
\def\vz{\mat{z}}
\def\vt{\mat{t}}
\def\vh{\mat{h}}
\def\mA{\mat{A}}
\def\mD{\mat{D}}
\def\sC{\set{C}}
\def\sH{\set{H}}
\def\sX{\set{X}}
\def\sS{\set{S}}
\def\sL{\mathbf{Labeled}}
\def\dvrd{\mathbf{D_\mathrm{vrd}}}
\def\imag{\mat{I}}
\def\conv{\circledast}

\def\csnl{CSNL}
\def\L{\mathcal{L}}
\def\C{\mathcal{C}}
\def\A{\mathfrak{A}}
\def\I{\mathcal{I}}
\def\R{\mathcal{R}}
\def\Y{\mathcal{Y}}
\def\D{\mathfrak{D}}
\renewcommand{\implies}{\rightarrow}
\def\loss{\mathcal{L}}
\newcommand{\logit}{\mathrm{logit}}
\def\digit{\mathrm{digit}}
\def\three{\mathit{three}}
\def\softselect{\mathit{softselect}}
\def\samp{\mathcal{S}}
\def\ie{\textit{i.e.}}
\def\eg{\textit{e.g.}}
\def\TRUE{\emph{True}}
\def\FALSE{\textit{False}}
\def\sT{\set{T}}
\def\sA{\set{A}}
\def\canride{\mathrm{CanRide}}
\def\riding{\mathrm{Riding}}
\def\vrd{\mathrm{vrd}}
\def\isridable{\mathrm{Ridable}}
\def\isliving{\mathrm{Living}}
\def\iswearable{\mathrm{Wearable}}
\def\wear{\mathrm{Wear}}

\definecolor{redcol}{rgb}{1, 0, 0}
\definecolor{bluecol}{rgb}{0, 0, 1}
\newcommand{\red}[1]{\textcolor{redcol}{#1}} 
\newcommand{\blue}[1]{\textcolor{bluecol}{#1}} 
\renewcommand{\paragraph}[1]{\smallskip\noindent{\bf{#1}}}

\def\algorithmautorefname{Algorithm}
\def\figureautorefname{Figure}
\def\tableautorefname{Table}
\def\equationautorefname{Eq.}
\def\sectionautorefname{Section}

\maketitle

\begin{abstract}
    We improve zero-shot learning (ZSL) by incorporating common-sense knowledge in DNNs.  
    We propose Common-Sense based Neuro-Symbolic Loss (\csnl) that formulates prior knowledge as novel neuro-symbolic loss functions that regularize visual-semantic embedding (VSE). \csnl{} 
    forces visual features in the VSE to obey common-sense rules relating to hypernyms and attributes. 
    We introduce two key novelties for improved learning-- (1) enforcement of rules for a group instead of a single concept to take into account class-wise relationships, and (2) confidence margins inside logical operators that enable implicit curriculum learning and prevent premature overfitting 
    We evaluate the advantages of incorporating each knowledge source and show consistent gains over prior state-of-art methods in both conventional and generalized ZSL \eg{} $11.5\%$, $+5.5\%$, and $11.6\%$ improvements on AWA2, CUB, and Kinetics respectively.
\end{abstract}

\section{Introduction}
\label{sec:intro}

\begin{figure*}[tbp!]
    \centering{\includegraphics[width=.6\textwidth]{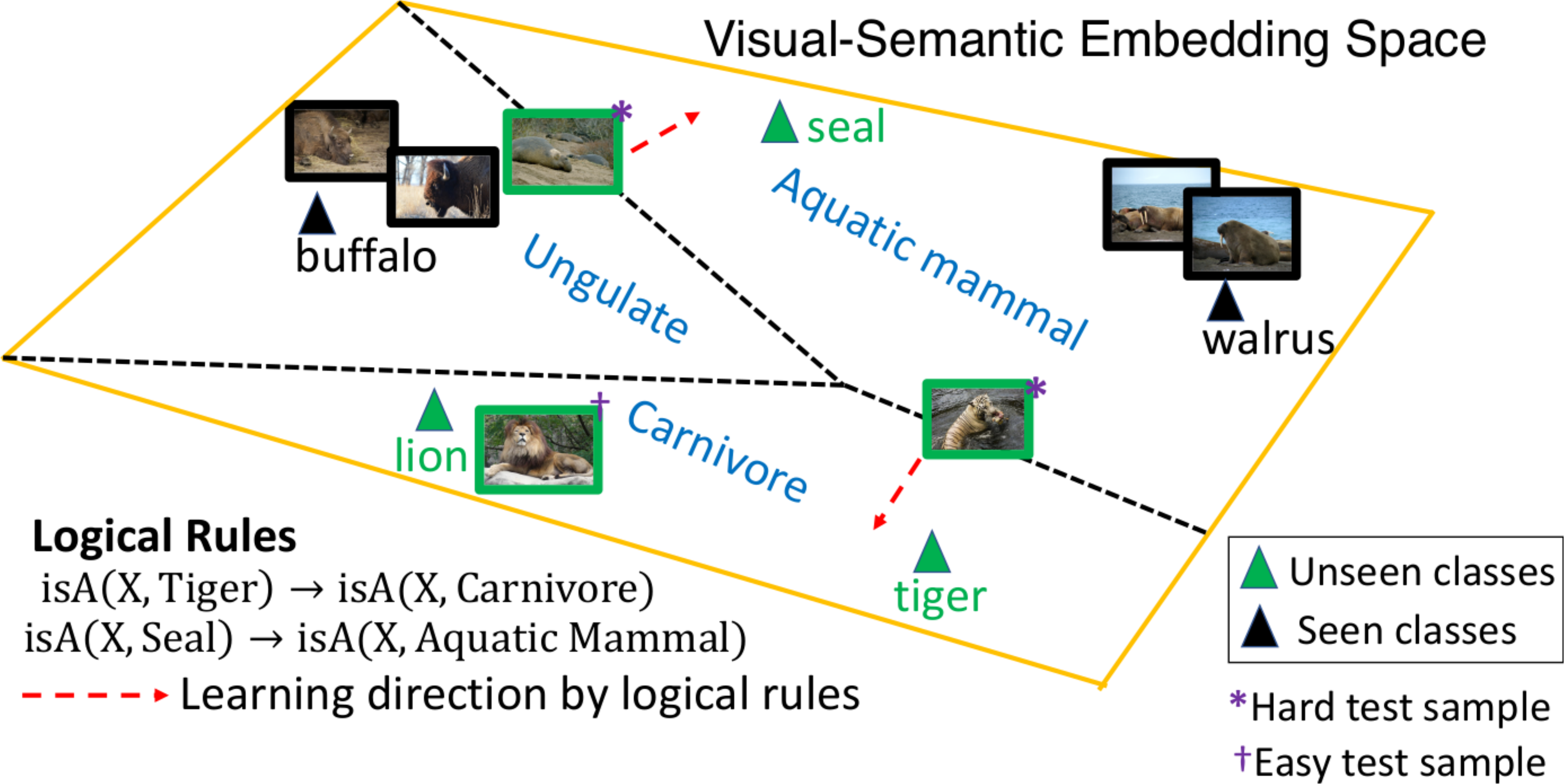}}
    \caption{
    Our proposed logical loss regularizes the embedding space by enforcing logical rules on visual samples. For example,  
    the rule \colons{$\text{isA}(\vx, \text{Tiger}) \rightarrow \text{isA}(\vx, \text{Carnivore})$} 
    leads to correct classification for the (hard) unseen test example by pushing it closer to its hypernym class-- \colons{Carnivore}, which occurs near \colons{Tiger} class. This improves ZSL by improving discriminability of the embedding space.}  
    \label{fig:main}
\end{figure*}

It is often assumed that the training data for visual recognition tasks has sufficient examples to cover all classes.
\cite{deng2009imagenet,lin2014microsoft}. However, this is unrealistic in real-world settings since the natural class distribution in most problems is heavy-tailed and thus many classes will only have a few samples \cite{wang2017learning, van2018inaturalist, sun2017revisiting}. Recent works have attempted to address these data scarce conditions by focusing on low shot settings, where the test classes either have a \textit{few} or \textit{zero} training examples \cite{xian2017zero,akata2015label, sung2018learning, hariharan2017low, snell2017prototypical, lampert2013attribute}. We focus on zero-shot learning (ZSL), where the model is evaluated on examples from (unseen) classes not available during training.
\textit{Visual-Semantic Embedding} (VSE) has emerged as one of the key methods for ZSL, which performs classification by embedding the visual features and the output labels in the same vector space.
Despite their success, such embeddings often fail to learn important class-level distinctions, especially for unseen classes, as they are learned from a 
selectively sampled subset of the visual space \cite{barz2019hierarchy, li2017learning}. This limits the  
generalizability of VSE in ZSL.  
We propose a novel \textit{neuro-symbolic loss} to address this issue. Our loss acts as a regularizer to improve the separability of classes in the embedding space by forcing visual features to obey logical rules that are written to express common-sense relationships (\autoref{fig:main}).

VSE projects visual features and output labels into a common space such that correct image-label pairs are closer based on some metric \cite{akata2015label,song2018transductive,frome2013devise, bansal2018zero}.
A test sample is classified by selecting the nearest label in the embedding space. 
Zero-shot classification is then performed by using label embeddings that contain both seen and unseen classes. 
In this work we focus on using pre-trained word embeddings for label embeddings since they do not require manual effort as compared to attributes \cite{joulin2016fasttext, le2014distributed, lampert2013attribute, akata2015label}. 
There have been several recent attempts to improve VSE 
by proposing novel loss functions, better embedding strategies or distance metrics, exploiting unlabeled data, etc. 
Despite these innovations, VSE maintains low performance on unseen classes as the learned space lacks the discriminative power to effectively separate such classes.  
This issue can also be understood under the lens of domain adaptation, where a classifier trained on a source domain (seen classes) fails to generalize to a novel target domain (unseen classes) \cite{wang2019unifying, kodirov2015unsupervised}. Another issue that plagues VSE is that the embedding space retains the artifacts of the original word embeddings that are often inconsistent with visual semantics and thus adversely affects class separability \cite{qiao2017visually}. For example, visual clusters corresponding to \colons{Cats} might be closer to \colons{Dogs} instead of semantically similar categories such as \colons{Tiger} and \colons{Lion}. Prior works have attempted to tackle these issues by incorporating class-level relationships \eg{} hypernyms in either the label embeddings or the loss function \cite{akata2015evaluation, miller1998wordnet, barz2019hierarchy, li2017learning}. However, these approaches require complex changes to the formulation of baseline VSE methods (\eg{} Devise \cite{frome2013devise}) and are still not able to address the above issues. 
We propose a simpler approach that preserves the formulation of the baseline VSE but improves performance by enforcing common-sense rules on visual samples in the embedding space. 


The proposed method, referred to as \textbf{Common-Sense based Neuro-Symbolic Loss} (\csnl), uses a novel neuro-symbolic loss function that enforces human-interpretable logical rules on a visual feature ($\vx \in \R^D$) such as "$\text{isA}(\vx, \text{Tiger}) \rightarrow \text{isA}(\vx, \text{Carnivore})$", where the value of the predicate \colons{$\text{isA}$} is computed using the VSE metric\footnote{This rule can be read as if $\vx$ is a \colons{Tiger}, then it is also a \colons{Carnivore}.}. \csnl{} leverages these relationships to semantically organize the VSE, which improves the discriminative ability of the embedding space, especially in under-sampled regions corresponding to unseen classes.  
We achieve this by extending a recently introduced neuro-symbolic method- Deep Adaptive Semantic Logic (DASL)- to translate the logical rules into a loss function \cite{sikka2020deep}. Compared to prior works that implement logical rules with simple implications \cite{sikka2020deep, donadello2017logic}, as presented above, we introduce two key novelties to support more effective learning. First, instead of enforcing rules for an individual concept, we enforce them at a \textit{set} level. As per this set-level rule, if a sample has high similarity to a hypernym class, then it should have high similarity to the member classes of that hypernym set.  
This formulation results in an improved embedding function 
by progressively learning representations for a hypernym set (\eg{} \colons{Feline}) and then for   
individual classes (\colons{Tiger} and \colons{Cats}). 
This prevents semantically implausible predictions for unseen classes that lie in parts of the embedding space with low classifier confidence.   
However, direct application of the \csnl{} is overwhelmed by overfitting caused by premature learning of rules. Our second contribution is therefore to modify logical implications to enable
prioritized enforcement of rules from simple to difficult examples, so that each rule is enforced only when there is sufficient confidence in its applicability to the example at hand. We implement this by adding tunable confidence margins within the \csnl{}, and adapting them such that backpropagation through the \csnl{} will automatically implement the desired “simple to difficult” curriculum.
Both these contributions also enable semi-supervised learning in ZSL, referred to as \textit{transductive learning}, where we are able to leverage unlabeled test data to improve performance. Although in this work we consider knowledge from hypernyms ("isA") and attributes ("hasA"), our method can handle multiple knowledge sources by tuning hyper-parameters for the losses induced by each knowledge source.
We evaluate our method on two standard zero-shot image classification datasets-- AwA2 and CUB,    
in both conventional and generalized settings. We also evaluate our model on zero-shot action classification on a proposed subset from the Kinetics dataset.   
We perform systematic studies to show the benefits of adding high-level knowledge from hypernyms and attributes, which results in consistent improvements over State-of-the-Art (SOTA) methods. Our contributions are:

\begin{enumerate}

    \item Common-Sense based Neuro-Symbolic Loss (\csnl) to regularize visual-semantic embedding by enforcing logical rules on visual features
    \item Advance over prior neuro-symbolic methods through two key novelties-- (1) enforcing rules for a group instead of an individual class to take into account class-level relationships, (2) implementing implicit curriculum learning within \csnl{} to prevent premature overfitting to logical rules
    \item Detailed ablation study to evaluate the benefits of hypernym and attribute based common-sense rules 
    \item SOTA results on zero-shot image and action classification tasks in both conventional and generalized settings.
\end{enumerate}

\section{Related Works}

\paragraph{Knowledge integration} in machine learning has been shown to be effective in cases of insufficient training data.
\cite{fung2003knowledge, mangasarian2007nonlinear} formulate prior knowledge as constraints and solve constrained optimizations to incorporate knowledge. 
\cite{jiang2018hybrid, liang2018symbolic} encode knowledge as graphs and explicitly integrate it through graph neural networks.
\cite{barz2019hierarchy} constrains learned feature similarities to approximate similarities from class hierarchy. \cite{verma2012learning} constructs a hierarchical model to explicitly integrate class taxonomy.
Our work is related to neuro-symbolic approaches that describe knowledge using logical rules, which are translated into loss functions to train neural networks \cite{diligenti2017integrating, hu2016harnessing, hu2016deep, stewart2017label, xu2018semantic, serafini2016logic, sikka2020deep}. These methods are prone to overfitting since they entail solving a hard optimization objective in presence of multiple rules. We tackle these issues by proposing two key novelties-- enforcing rules at a set level that takes into account class-level similarities and prevents catastrophic failures, and an implicit curriculum based approach for prioritized enforcement of rules to prevent premature overfitting to rules.

\paragraph{Zero-shot learning}
Early methods utilize intermediate representations to connect seen and unseen classes. \cite{al2016recovering, jayaraman2014zero, kankuekul2012online}. \cite{lampert2013attribute} learns to predict attributes from input images and then predict class labels based on these attributes. \cite{norouzi2013zero} instead uses seen class posteriors to predict unseen labels. 
These two-stage methods suffer from a weak intermediate representation that is unable to generalize well on unseen classes. 
Recent methods have focused on learning a common embedding space, referred to as a Visual Semantic Embedding (VSE), by projecting both input features and class labels \cite{frome2013devise,akata2015label}. 
The works have explored class embeddings based on both attribute and (pre-trained) word embeddings.
Although attribute embeddings outperform word embeddings, they are difficult to obtain in real-world applications.   
\cite{frome2013devise} learns a mapping from visual features to semantic features, and \cite{zhang2017learning} proposes mapping from semantic features to visual features. In order to make the joint embedding better generalizable to unseen data, \cite{akata2015evaluation} augments the attribute embeddings with hierarchical embeddings from WordNet. \cite{xian2016latent} further improves this mapping function by making it nonlinear.  
Despite these advancements, the learned embedding space is heavily biased towards seen classes and is unable to effectively separate unseen classes.
We improve VSE by regularizing the embedding with a logical loss that enforces common-sense rules over the input features.   


To investigate the performance of ZSL models under a more generic setting, generalized zero-shot learning (GZSL) was proposed in \cite{scheirer2012toward} 
where both seen and unseen classes are present at test time. \cite{chao2016empirical} and \cite{xian2017zero} show that models 
achieving high ZSL performance do not necessarily perform well under the GZSL since the predictions are 
highly biased towards seen classes. \cite{song2018transductive} alleviates this problem by introducing a transductive loss that guides 
predictions for unlabeled test data towards the unseen classes. We also show the benefits of our loss in the GZSL. 

Recent works have also used feature generative models \cite{xian2018feature, zhu2018generative, huang2019generative, li2019leveraging, zhu2019learning, sariyildiz2019gradient}. These methods use generative networks to synthesize inputs/features in unseen classes, and thus turn ZSL problems into normal classification problems. 

\paragraph{Few-shot learning} implicitly integrates prior knowledge by learning from similar tasks.
\cite{benaim2018one, hu2018few, motiian2017few, zhang2018fine} learn models for multiple tasks with earlier layers shared among all models. \cite{yan2015multi, luo2017label} instead encourage the models' parameters to be similar. 
Recently, meta-learning methods \cite{vinyals2016matching, snell2017prototypical, shyam2017attentive, sung2018learning} learn a common 
embedding space from large-scale datasets for similar tasks, which generalizes well to few-shot tasks.

\paragraph{Semi-supervised learning}
\cite{triguero2015self, zhou2010semi, chen2016xgboost} alternate between training and pseudo-labelling to learn models with both labelled and unlabelled data. 
\cite{sheikhpour2017survey, goldberg2009multi, dara2002clustering} perform unsupervised feature extraction or clustering to obtain latent representations and then learn prediction models from supervised data. 
\cite{salimans2016improved, odena2016semi, kingma2014semi} train generative models to learn better latent representations. Our work also performs semi-supervised learning by instead enforcing common-sense rules on unlabeled test data that is supported by an implicit curriculum to prevent overfitting. 

\section{Approach}
Visual Semantic Embedding (VSE) performs zero-shot learning (ZSL) by projecting visual features and class embedding into a common space where test samples are classified into unseen classes using its nearest neighbor \cite{akata2015label,frome2013devise}. However, as discussed in \autoref{sec:intro}, such embeddings lack the discriminative ability to generalize well to unseen classes. We address this problem by proposing a novel loss function that forces visual features within the embedding space to obey common-sense logical rules.  
This loss results in an improved semantic organization of the embedding space by exploiting class-level relationships, such as hypernyms, and leads to improved performance on unseen classes. 
We now describe our approach. It consists of a baseline VSE model and the proposed neuro-symbolic loss function. 

\subsection{Baseline VSE for Zero-Shot Learning}
\label{sec:baseline}

We denote the training dataset as $\mathcal{D}^s = \{(\vx_i^s, y_i^s)\}_{i=1}^{N_s}$, where each visual feature $\vx_i^s \in \R^{D_x}$ is associated with a corresponding label $y_i^s \in \mathcal{Y}^s$, and $\mathcal{Y}^s$ is the set of seen classes. We denote the test  dataset $\mathcal{D}^t = \{(\vx_i^t, y_i^t)\}_{i=1}^{N_t}$, where $y_i^t \in \mathcal{Y}^t$, and $\mathcal{Y}^t$ is set of unseen classes. There is no overlap between the set of seen and unseen classes $\mathcal{Y}^s \cap \mathcal{Y}^t = \emptyset$, $\mathcal{Y}^s \cup \mathcal{Y}^t = \mathcal{Y}$. We denote the label embedding of class $y$ as $\ve_y \in \R^{D_y}$, which is obtained from pre-trained word embeddings. 
Our goal is to generalize a classifier learned from seen class samples to unseen class samples. In the \textit{conventional} ZSL, the test samples belong to only unseen classes, while in the \textit{generalized} ZSL (GSZL), the test samples can belong to both seen and unseen classes \cite{xian2017zero}. To do well on both settings, we operate
in a transductive setting which assumes that we are provided with the labeled seen dataset $\mathcal{D}^s$, the unlabeled test dataset $\mathcal{D}^{\text{trans}} = \{x_i^t\}_{i=1}^{N_t}$ and the label embedding of all the classes during training. 
We use the baseline embedding model from \cite{song2018transductive}, which is based a common VSE method-- Devise
\cite{frome2013devise}. 
This baseline model projects $D_x$ dimensional visual features and $D_y$ dimensional textual features into a common $D_e$ dimensional embedding space using linear projections. The conditional probability of an example $\vx$ lying in class $y$ is computed using cosine similarity with a $softmax$ function:
\begin{align}
    s(\vx, y) &= \frac{(W_x^Tx)^T (W_y^T\ve_y)}{\norm{(W_x^T\vx)}\norm{(W_y^T\ve_y)}}  \\
    p(y|x) &= \frac{e^{\gamma s(x, y)}}{\sum_{y \in \Y} e^{\gamma s(x, y)}} 
    \label{eq:scoring_fn}
\end{align}
where $s(\vx,y)$ and $p(y|x)$ are the scores and probability respectively for sample $\vx$ for class $y$, $W_x \in \R^{D_x \times D_e}$ and $W_y \in \R^{D_y \times D_e}$ are projection matrices, and $\gamma$ is a constant multiplier. Similar to \cite{song2018transductive}, we perform a $softmax$ over all the classes ($\Y$) during training. The classification loss $\L_c(x)$ is computed using the cross-entropy loss function.

Following \cite{song2018transductive}, we add an additional loss term $\L_q$ that reduces bias from the seen classes by increasing the sum of probabilities of unlabeled test examples for the unseen classes. 
This loss and the final training loss are computed as
\begin{align}
    &\L_q(x) = -\ln\sum_{y\in\mathcal\mathcal{Y}^t} p(y|x)  \\ 
    \label{eq:loss}
    &\L_t = \frac{1}{N_s}\sum_{i=1}^{N_s} \L_c(\vx_i^s) +  \frac{\lambda_q}{N_t}\sum_{i=1}^{N_t} \L_q(x_i^t) + \lambda_{reg} \Psi(W)  
\end{align}
where $\Psi$ is an $l^2$-norm regularizer, and $\lambda_q$ and $\lambda_{reg}$ are regularization parameters.


\subsection{Common-Sense based Neuro-Symbolic Loss (\csnl)}
\label{sec:cnsl}

We incorporate common-sense rules to regularize the above embedding method by adding the loss terms from \csnl{} to \autoref{eq:loss}.    
\csnl{} enforces logical rules that visual features must obey in the common embedding space. For example, we know that \colons{Tigers} are \colons{Carnivore}, so any image labeled as a \colons{Tiger} must also be labeled as a \colons{Carnivore} (see \autoref{fig:main}). These additional rules are meant to regularize the embedding space and are often superfluous if sufficient labeled data is available, but provide utility when working in a low data regime such as ZSL. We implement these logical rules using a recent neuro-symbolic method-- Deep Adaptive Semantic Logic (DASL) \cite{sikka2020deep}-- that integrates user-provided knowledge in first order logic (FOL) with training data to learn neural networks. DASL represents truth values in pseudo-probabilities denoted as $t$ to support differentiation. $t$ is implemented as a standard sigmoid function such that $\logit(t(x)) = x$. We express logical rules using DASL's modified FOL formalism, then use DASL to compile a neural network layer (as a \textit{torch.nn.Module}) that implements these rules \cite{paszke2019pytorch}. 
This results in multiple binary outputs $z_i$ (represented as logits in the range $[-\infty, \infty]$), one for each logical assertion. 
We then add a logical loss to the original loss function in \autoref{eq:loss} as
\begin{equation}
    \L = \L_t + \sum_i \lambda_i^{L}\text{bce}(z_i, 1.0)
    \label{eq:logic_loss}
\end{equation}
where $\text{bce}()$ is the binary cross entropy with logits function, and $\lambda_i^{L}$ is the regularizer for the $i^{th}$ loss term. 

We use two common-sense rules to provide semantic regularization-- a \textbf{hypernym rule} that describes an expected hierarchy among the classes and an \textbf{attribute rule} that describes low level binary features that should be present or absent conditional upon the object class. A hypernym rule is formulated as
\begin{equation}
    (\forall \vx \in \sX) (\forall \vh \in \sH) [\text{isA}(\vx, \vh) \implies \text{inside}(\vx, \sC_h)]
    \label{eq:hyper_logic}
\end{equation}
Here we use sorted logic, where $\sX$ is the set of images/videos, and $\sH$ is a set of hypernyms. $\sC_h$ is the subset of classes consistent with hypernym $h$, \colons{isA} is a learned neural network, and \colons{inside} is a function that returns True if the image $\vx$ is predicted to belong to one of the classes in $\sC_h$. For binary attributes $a$ we have a similar equation
\begin{equation}
    (\forall \vx \in \sX) [\text{isA}(\vx, a) \iff \text{inside}(\vx, \sC_a)]
    \label{eq:attr_logic}
\end{equation}
The notation is the same as in \autoref{eq:hyper_logic}, except $a$ is an attribute, and $\sC_a$ is the set of classes consistent with that attribute. 
While the above equations are for a single hypernym set $\sH$ and a single attribute $a$ respectively, extension to multiple hypernym sets and multiple attributes is straightforward.
The \colons{isA} predicate is realized using the learned scoring function in \autoref{eq:scoring_fn} to allow the logical rules to backpropagate and influence the embedding space. We note that this logical rule is different from the example shown in \autoref{fig:main}, which focuses on enforcing rules for a single class. The advantage of using the proposed \colons{set level} versus a \colons{concept level} logical formulation is that it 
guides the embedding space to progressively learn representations for a set \eg{} \colons{Carnivore} hypernym, followed by individual classes. This prevents catastrophic failures for unseen classes by effectively partitioning the embedding space (see \autoref{fig:main}).

To support better learning and integration with the baseline ZSL method, we modify the implementation of \autoref{eq:hyper_logic} and \autoref{eq:attr_logic} in two ways. The first modification is to adjust the set membership function above from its standard FOL implementation to a version that works with a $softmax$ neural network implementation, as output by the base embedding in \autoref{eq:scoring_fn}. The second modification is to introduce a \textit{confidence margin} into the logical implication operators to help the neural network avoid local minima induced by the logical rules.
Using DASL's modified first order logic we could directly implement set membership as
\begin{equation}
    \text{inside}(\vx, \sS) = (\exists c \in \sS) [\text{isA}(\vx, c)]
    \label{eq:inside_simple}
\end{equation}
where $c$ is a class label and $\sS$ is the hypernym/attribute set. \autoref{eq:inside_simple} will return True if the model is confident (has high truth value) that the object $\vx$ is a specific element of the set. 
However, this formulation does not correctly account for the fact that the classes are mutually exclusive. We therefore use the more mathematically precise formulation: 
\begin{align}
    \text{inside}(\vx, \sS) &= N_s \sum_{y \in S} p(y|x)
\end{align}
which is True if the truth values are well localized within the set, even if we are unsure which set member was actually detected. For example, it still correctly predicts \colons{Feline} even if we are uncertain about the object being a \colons{Cat} or a \colons{Tiger}, which is important to minimize catastrophic failure for examples from the unseen classes. We normalize with a constant $N_s$ such that the truth value is $\frac{1}{2}$ for a uniform distribution over classes. Both \colons{$\text{inside}$} and \colons{$\text{isA}$} are converted into DASL's logit representation \cite{sikka2020deep} (using $\logit$ function). 

Logical rules, if implemented directly, often have the unfortunate consequence of introducing undesired local minima into the optimization landscape. Consider the simple rule $a \iff b$, that requires both $a$ and $b$ to have the same truth value. Using the product t-norm to implement this rule as $(\neg a \vee b) \wedge (\neg b \vee a)$ there are two local minima, as expected, with $t(a) = t(b) = 1.0$ or $t(a) = t(b) = 0.0$. If we are initially uncertain about both truth values (e.g. $t(a) = t(b) = .5$), which is usually the case, then small perturbations will select one of the local minima. More concretely, if we start with a small deviation such that $t(a) = t(b) = .5 + \epsilon$, then backpropagation will iteratively reinforce this small initial bias, quickly getting trapped in a local minimum at $t(a) = t(b) = 1.0$. This minimum does correctly solve the rule; however if we later learn that $t(a) = 0.0$, it will be too late to correct to the minimum at $t(a) = t(b) = 0.0$. 
In the presence of multiple rules, data points, and random network initialization, this effect will lead to a large number of early \colons{decisions} by the neural network and thus trapping the neural network in a deep local minimum of the logical regularizer 
before it has the opportunity to effectively learn the data.

To avoid these local minima we need an algorithm that delays decisions about how to solve rules until there is sufficient information to make these decisions correctly. Specifically, we must have prioritized enforcement of rules, such that simple cases are enforced first while more ambiguous cases are enforced later. 
This is similar to using curriculum learning \cite{bengio2009curriculum}, but without the need for manual creation of a curriculum. We achieve this by modifying our logical loss such that backpropagation over the modified loss will emulate the Unit Claus Propagation (UCP) algorithm used to solve simple satisfiability problems. UCP iteratively assigns values to unknown variables based on enforcing rules in priority order, where traditionally priority for rules is determined by the number of free variables in a clause \cite{chao1986probabilistic, achlioptas2001lower}. To emulate this we modify the disjunction operator to include a confidence margin ($c$). The margin is added to the inputs of the disjunction, and subtracted from the output ($((a+c) \vee (b+c))-c$) ($a$, $b$ and $a \vee b$ are represented in DASL's logit representation).
For cases where all disjuncts have roughly equal truth value the overall effect is to increase the truth value of the input by $(n-1)*c$, where $n$ is the number of uncertain disjuncts. 
In implication this means that if the certainty about both the consequent and antecedent is low (truth value $\leq c$), then that biases the output towards True and thus the loss will not depend on that implication. 
However, if we are confident either about the consequent or antecedent (truth value $>>c$), then the implication will be enforced by the loss function during backpropagation as if there were no confidence margins. For example, this will enforce the rule $\text{isA}(\vx, \text{Tiger}) \rightarrow \text{isA}(\vx, \text{Feline})$ only when the model is able to confidently make predictions either about \colons{Tiger} or \colons{Feline}. 
As certainty over variables increases during training (due to improving NN weights), the backpropagation enforces more rules. We treat the confidence margin as a hyperparameter of the optimization.


Instead of fixing a single confidence margin we linearly sweep the confidence margin, starting at a high value, $c_\text{start}$, and moving to a low value, $c_\text{stop}$, over a fixed number of training epochs, $c_\text{epochs}$. 
This enables a dynamic curriculum that enforces more rules as the training proceeds and the underlying network becomes more reliable. 
This is particularly useful for enforcing rules on the unlabeled test data in the transductive setting, which could easily overfit on the rules. 
The final loss used in our experiments is expressed as:
\begin{align}
    \L &= \L_t + \L_{\text{\csnl}} (\mathcal{D}^s) + \lambda_{\text{trans}}\L_{\text{\csnl}} (\mathcal{D}^{\text{trans}})  \\
    \L_{\text{\csnl}}(\mathcal{D}) &= \sum_{i=1}^{N} (\lambda_{\text{hyp}} L_{\text{hyp}}(x_i) + \lambda_{\text{attr}} L_{\text{attr}}(x_i)) 
\end{align}
where $L_{\text{hyp}}$ and $L_{\text{attr}}$ are the \csnl{} based loss functions for hypernym and attribute rules respectively, 
$\mathcal{D}=\{x_i\}_{i=1}^{N}$ denotes a dataset with $N$ samples, $\mathcal{D}^s$ and $\mathcal{D}^{\text{trans}}$ were introduced in \autoref{sec:baseline}, $\lambda_{\text{hyp}}$, $\lambda_{\text{attr}}$, and $\lambda_{\text{trans}}$ are regularization parameters. 
$\L_{\text{\csnl}} (\mathcal{D}^{\text{trans}})$ is used during transductive setting to enforce rules ($\lambda_{\text{trans}}=0$ during conventional ZSL).   



\section{Experiments}
\label{sec:experiments}

\subsection{Overview}
We evaluate our approach on zero-shot image and action classification. 
We briefly describe the datasets, the metrics, and the implementation details. We then provide quantitative results that include an ablation study to show the contribution of different common-sense rules enforced using our loss, and a study of confidence margin parameters. 
We finally compare our model with SOTA methods.

\paragraph{Datasets:} We perform ZSL experiments on two standard image classification datasets: AWA2 and CUB \cite{xian2017zero}, and introduce a new subset from the Kinetics dataset for zero-shot action classification. 
 \textbf{AWA2} comprises of $37,322$ images from $50$ coarse-grained animal classes, with each class annotated with $85$ shared attributes describing the color, body shape, habitat \etc. \textbf{CUB} \cite{wah2011caltech} contains $11,788$ images from $200$ bird species, with images annotated with $312$ binary attributes. AWA2 has $40$ seen and $10$ unseen classes, while CUB has $150$ seen and $50$ unseen classes.  
 We use image embeddings (ResNet-101 pre-trained on ImageNet), attributes and data splits made available by Xian \etal \cite{xian2017zero}.

To evaluate our approach on zero-shot action classification we propose a new subset from the Kinetics dataset \cite{kay2017kinetics}, referred to as \textbf{Kinetics-ZS}. 
We focus on a set of fine-grained sports classes (see appendix for more details) since such classes would require a discriminative embedding space that can encode distinct motion and appearance patterns.
This allows us to study the efficacy of the proposed loss in enforcing common-sense rules and improving upon pre-trained spatio-temporal features (SlowFast network \cite{feichtenhofer2018slowfast}). 
We selected $91$ seen classes from the Kinetics-400 dataset \cite{kay2017kinetics} that contains around $300$K videos from $400$ human action classes.  
We then use Kinetics-600 dataset \cite{carreira2018short} to choose $18$ unseen classes, which have no overlap with the classes from the Kinetics-400 dataset.
We do this to fairly use the video features from the SlowFast network for ZSL, which is pre-trained on Kinetics 400.  The final dataset contained $3861$ videos from $91$ seen and $18$ unseen classes\footnote{We will release the splits soon.}. We have provided additional details in \autoref{sec:kinetics} in the supplementary.



\paragraph{Evaluation Metrics:} Following prior works \cite{xian2017zero}, we use the mean class accuracy (MCA) metric for evaluation:
\begin{align}
    MCA = \frac{1}{|\mathcal{Y}|}\sum_{y\in \mathcal{Y}} acc_y,
\end{align}
where $acc_y$ is the top-1 classification accuracy for class $y$. In the conventional ZSL, both the test data and the prediction space of the classifier is restricted to unseen classes ($\text{MCA}_t$). However, in the generalized setting (GZSL), both the test data and the search space of the classifier includes seen and unseen classes. We follow prior works \cite{xian2017zero} and report harmonic mean ($\text{HM}$) of the MCA for both seen ($MCA^g_s$) and unseen classes ($MCA^g_t$):
\begin{equation}
    \text{HM} = \frac{2 * MCA^g_s * MCA^g_t}{MCA^g_s + MCA^g_t}
    \label{eq:gzsl}
\end{equation}

\paragraph{Implementation Details:}
We first map each class in the dataset to its WordNet synset \cite{miller1998wordnet}. 
We obtain hypernyms for each dataset class by first selecting a root synset whose subtree contains a diverse set of classes and 
then defining its immediate children (using WordNet) as hypernym classes. We declare a class to be consistent with a hypernym class if its associated synset is a descendant of that hypernym (used to create hypernym set in \autoref{eq:hyper_logic}). 
For example, for AWA2 we use \colons{Placental.n.01} as the root synset (see \autoref{fig:class_hierarchy} in supplementary) whose children create hypernym categories such as \colons{Carnivore} and \colons{Aquatic-mammal} that are able to provide decent coverage for all the classes in this dataset.
Although this strategy reduces the WordNet tree to a two-level hierarchy, it provides useful information while keeping the complexity minimal. Similarly, we obtain the hypernyms for CUB classes from the taxonomy provided in \cite{morgado2017semantically}, which is also based on WordNet. We manually select a subset of hypernym classes such that they cover the CUB classes in a balanced manner. For Kinetics-ZS, we derive hypernyms based on the proposed list of parent-child groupings provided in \cite{kay2017kinetics}. 
We obtain $300$ dimensional class embeddings for the dataset and hypernym classes from the FastText model trained on Wikipedia \cite{joulin2016fasttext}. 
If a hypernym class does not exist in the FastText model, we replace it with the average embedding of its components (\eg{} aquatic + mammal) or synonyms/children nodes. For AWA2, we use the binary class-level attributes provided by the authors. 
We derive class-level attributes for CUB by binarizing the averaged image-level attributes with a threshold of $0.75$. We manually labeled $20$ attributes for Kinetics.  
We refer the reader to \autoref{table:kinetics_classes} and \autoref{table:kinetics_attributes} in supplementary for the list of hypernyms and attributes for Kinetics-ZS.
We use linear layers to project word embeddings and image features into a $1024$ dimensional common embedding space. 
Since PySlowFast was pre-trained on Kinetics-400, we also add randomly generated Gaussian noise ($\mu=0$, $\sigma=0.5$) to the image features prior to projection. We do this to shift the image vectors and loosen the tight clusterings resulting from feature extraction. 
Following prior works, we only learn the projection layer during training \cite{xian2017zero,song2018transductive,frome2013devise}. 
We select hyperparameters such as learning rate, confidence start, stop and epochs, and regularization parameters using the validation splits provided by Xian et al. The $\gamma$ in \autoref{eq:scoring_fn} is set to $32$. We use the Adam optimizer with a weight decay of $10^{-5}$ and set batch-size to $128$. 
All the implemented SOTA methods use the same experimental settings (\eg{} splits, features) as ours for a fair comparison (see \autoref{sec:implementation} in supplementary).

\subsection{Quantitative Result}

\begin{table*}[h!]
    \centering
    \begin{tabular}{c|ccc|cc|cc|cc}
        \hline
        Model & Hypernym & Attribute & Transductive & \multicolumn{2}{c|}{AWA2} & \multicolumn{2}{c|}{CUB} & \multicolumn{2}{c}{Kinetics-ZS} \\
         & rules & rules & learning & $\text{MCA}_t$ & $\text{HM}$ & $\text{MCA}_t$ & $\text{HM}$ & $\text{MCA}_t$ & $\text{HM}$ \\
        \hline
        Baseline & X & X & X & $47.3$ & $0.0$ & $15.9$ & $0.0$ & $42.8$ & $0.0$ \\ \hline
        \multirow{3}{*}{\textbf{\csnl}} & X & \checkmark & X & $59.6$ & $0.0$ & $29.3$ & $0.1$ & $48.1$ & $0.1$\\
         & \checkmark & X & X & $51.3$ & $0.0$ & $23.5$ & $0.1$ & $48.6$ & $0.1$ \\
         & \checkmark & \checkmark & X & $61.0$ & $0.0$ & $\mathbf{32.5}$ & $0.7$ & $50.2$ & $0.2$ \\ \hline 
        Baseline$^{\text{tr}}$& X & X & \checkmark & $49.3$ & $42.2$ & $14.9$ & $19.9$ & $55.9$ & $41.9$ \\ \hline
        \multirow{3}{*}{\textbf{\csnl$^{\textbf{tr}}$}} & X & \checkmark & \checkmark & $64.0$ & $55.8$ & $20.1$ & $24.6$ & $60.7$ & $44.2$ \\
         & \checkmark & X & \checkmark & $51.3$ & $58.1$ & $21.2$ & $26.7$ & $63.4$ & $45.4$  \\
         & \checkmark & \checkmark & \checkmark & $\mathbf{71.4}$ & $\mathbf{64.0}$ & $25.1$ & $\mathbf{29.4}$ & $\mathbf{67.5}$ & $\mathbf{47.0}$ \\
    \end{tabular}
    \caption{Ablation study showing performance with different logical rules enforced using \csnl{}. We report ZSL metrics on the conventional ($\text{MCA}_t$) and generalized setting ($\text{HM}$). $^{\text{tr}}$ superscript refers to operating in the transductive setting where the model has access to unlabeled test data. We see consistent improvements by incorporating different common-sense rules.} 
    \label{table:ablation_rules}
    \vspace{-1em}
\end{table*}

\subsubsection{Ablation Study}
\label{sec:ablation}
We study the advantages of adding common-sense rules based on hypernyms and attributes using CSNL in \autoref{table:ablation_rules}. 
We report both $\text{MCA}_t$ and the $\text{HM}$ metrics for evaluating ZSL in the conventional and the generalized settings respectively. 
We first focus on conventional ZSL without the use of unlabeled test data (transductive setting), which we later exploit to achieve further gains particularly on GZSL.  


Compared to the baseline ($\text{MCA}_t=47.3$ on AWA2, $15.9$ on CUB, $42.9$ on Kinetics-ZS), we observe consistent improvement with the hypernym rules ($51.3$ on AWA2, $23.5$ on CUB, $48.6$ on Kinetics-ZS) and the attribute rules ($59.6$ on AWA2, $29.3$ on CUB, $48.1$ on Kinetics-ZS) in the conventional ZSL. We see further improvements when combining the loss functions from both the attribute and hypernym rules ($61.0$ on AWA2, $32.5$ on CUB, $50.2$ on Kinetics-ZS). These results highlight the benefits of enforcing common-sense rules inside the embedding space.

In the transductive setting, the baseline (denoted as Baseline$^{\text{tr}}$) adds the additional loss term $\L_q$ (in \autoref{eq:loss}) \cite{song2018transductive} 
to avoid the bias towards the seen classes. 
Compared to Baseline$^{\text{tr}}$ ($\text{MCA}_t=49.3$ and $\text{HM}=42.2$ for AWA2), we see consistent improvements with the hypernym rules ($\text{MCA}_t=51.3$ and $\text{HM}=58.1$) and the attribute rules ($\text{MCA}_t=64.0$ and $\text{HM}=55.8$). Similarly to the conventional setting, the best performance on all datasets is achieved by combining the two logical rules \eg{} $\text{HM}=47.0$ for \csnl$^{\text{tr}}$ versus $41.9$ of Baseline$^{\text{tr}}$ on Kinetics-ZS.

Empirical gains on both metrics in the transductive setting  
show that \csnl{} is able to effectively leverage unlabeled data when adding common-sense knowledge.
We believe that the satisfaction of higher order relationships improves separability of classes lying in under-sampled regions of the embedding space,  particularly for unseen classes.  
We also note that the hypernym rules show better performance in the transductive setting on the GZSL metric compared to the attribute rules on all datasets ($26.7$ vs $24.6$ on CUB). We believe this happens because the hypernym rules can provide a more coherent partitioning of the embedding space than the attribute rules.

We also verify the contribution of the two novelties in \csnl-- set level rules and confidence margins-- by removing them from \csnl$^{\text{tr}}$. When replacing the set level rules with concept level rules, the performance on AWA2 drops from $\text{MCA}_t=71.4$ and $\text{HM}=64.0$ to $\text{MCA}_t=49.3$ and $\text{HM}=42.2$. Without the confidence margins in \csnl{}, the performance drop to $\text{MCA}_t=55.9$ and $\text{HM}=51.6$. 
The drop in performance corroborates the importance of our contributions over the baseline neuro-symbolic approach (DASL \cite{sikka2020deep} in achieving good performance.



\subsubsection{Study of Confidence Margin Parameters}
\label{sec:conf_margin}

\begin{figure}[t]
    \begin{subfigure}{.495\columnwidth}
      \includegraphics[width=\textwidth]{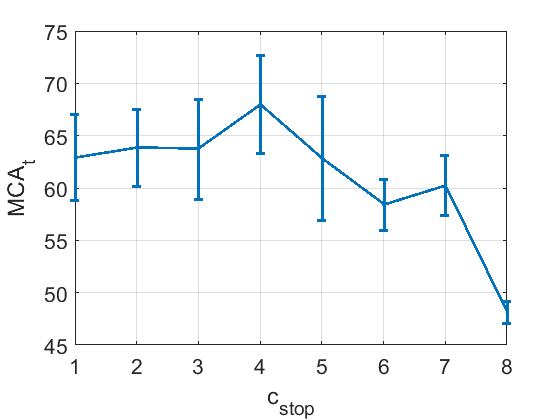}
      \vspace{-1em}
      \caption{}
      \label{fig:conf_awa-1}
    \end{subfigure}
    \begin{subfigure}{.495\columnwidth}
        \includegraphics[width=\textwidth]{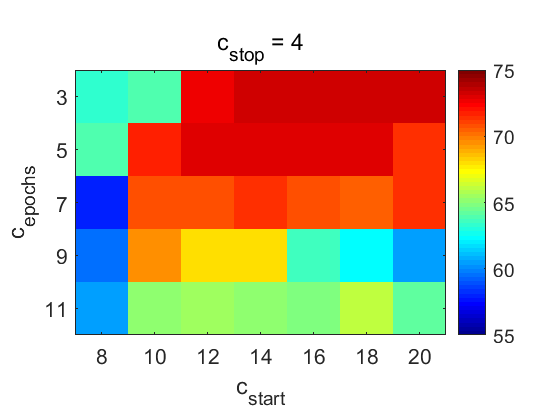}
        \vspace{-1em}
        \caption{}
        \label{fig:conf_awa-2}
      \end{subfigure}
    \label{fig:conf_awa}
    \caption{Effect of hyperparameters controlling the confidence margin in \csnl{} on zero-shot classification on AWA2. Left plot shows averaged performance for $c_\text{stop}$ while keeping $c_\text{start}$ and $c_\text{epochs}$ fixed. Right plot shows 2D plot for $c_\text{start}$ and $c_\text{epochs}$ while keeping $c_\text{stop}$ fixed (best seen in color).} 
\end{figure}

We introduced confidence margins in the logical loss function in \autoref{sec:cnsl} to progressively enforce rules from easy to difficult samples and prevent premature overfitting to the rules. We achieve this by linearly sweeping the confidence margin from a start value $c_\text{start}$ to a final value of $c_\text{stop}$ over $c_\text{epochs}$ training epochs. \autoref{fig:conf_awa-1} shows the sensitivity of the $\text{MCA}_t$ metric (on AWA2) to   
these parameters.
The left figure shows a plot of $\text{MCA}_t$ and  $c_\text{stop}$, with performance at each value of  $c_\text{stop}$ averaged across the other two hyperparameters. We observe that the performance falls to the baseline ($47.3$) for higher values of  $c_\text{stop}$ ($\ge 5$) since the logical rules will not be enforced due to the requirement of high confidence on predictions.  
However, for smaller values of  $c_\text{stop}$ ($\le 2$), the logical rules will be enforced prematurely resulting in a local minimum that satisfies the rules but disagrees with the data. For example, the model classifies all examples into \colons{Tiger} and all hypernyms into \colons{Carnivore}.

\autoref{fig:conf_awa-2} shows $\text{MCA}_t$ versus $c_\text{start}$ and $c_\text{epochs}$ at $c_\text{stop}=4$ (best value from the previous plot). We see a possible inverse relationship between $c_\text{start}$ and $c_\text{epochs}$ for achieving high performance \eg{} $\text{MCA}_t > 70$ for $(c_\text{start}=20, c_\text{stop}=3)$ and $(c_\text{start}=14, c_\text{stop}=5)$. This makes sense since if both $c_\text{start}$ and $c_\text{epochs}$ are high, then the rules will not be enforced completely (bottom right), while when both are low, rules will be enforced too quickly leading to overfitting. These curves highlight the need for confidence margins in \csnl{} to enable effectively learning. 


\begin{table*}[h!]
    \centering
    \begin{tabular}{c|cc|cc|cc}
        \hline
        Method & \multicolumn{2}{c|}{AWA2} & \multicolumn{2}{c|}{CUB} & \multicolumn{2}{c}{Kinetics-ZS} \\
         & $\text{MCA}_t$ & $\text{HM}$ & $\text{MCA}_t$ & $\text{HM}$ & $\text{MCA}_t$ & $\text{HM}$\\ \hline
        Zhang \cite{zhang2017learning} & $59.9$ & $35.7$ & $19.6$ & $6.2$ & $54.9$ & $23.1$ \\
        SJE \cite{akata2015evaluation} & $51.5$ & $0.0$  & $16.5$ & $0.0$ & $49.0$ & $0.0$  \\
        Latem \cite{xian2016latent}    & $52.5$ & $0.0$  & $19.7$ & $0.0$ & $51.0$ & $0.0$ \\
        Latem-Hier \cite{xian2016latent}  & $57.7$ & $0.0$  & $23.2$ & $0.0$ & $-$ & $-$ \\
        Devise \cite{frome2013devise}  & $51.7$ & $0.0$  & $18.7$ & $0.7$ & $51.9$ & $0.0$\\
        Devise (image space) \cite{frome2013devise} & $52.5$ & $0.0$ & $14.2$ & $0.0$ & $50.7$ & $0.0$ \\
        Conse \cite{norouzi2013zero}   & $47.9$ & $0.0$ & $17.4$ & $0.0$ & $49.6$ & $0.0$\\
        Learning to Compare \cite{sung2018learning} & $46.8$ & $28.1$ & $12.6$ & $8.8$ & $40.5$ & $26.4$\\
        QFSL$^{\text{tr}}$ \cite{song2018transductive} & $49.3$ & $42.2$ & $14.9$ & $19.9$ & $55.9$ & $41.9$\\
        \textbf{\csnl} & $61.0$ & $0.0$ & $\mathbf{32.5}$ & $0.7$ & $50.2$ & $0.2$ \\
        \textbf{\csnl}$^{\textbf{tr}}$ & $\mathbf{71.4}$ & $\mathbf{64.0}$ & $25.1$ & $\mathbf{29.4}$ & $\mathbf{67.5}$ & $\mathbf{47.0}$\\
    \end{tabular}
    \caption{Comparison of the proposed approach \csnl{} with SOTA methods on zero-shot image (AWA2, CUB) and action classification (Kinetics-ZS) tasks. We report metrics on conventional ($\text{MCA}_t$) and generalized settings ($\text{HM}$). $^{\text{tr}}$ superscript refers to operating in the transductive setting where the model has access to unlabeled test data.}
    \label{table:sota}
\end{table*}

\subsection{Comparison with SOTA}
\autoref{table:sota} shows that \csnl{} outperforms all SOTA methods in both the conventional and generalized settings on all datasets. 
For example, in comparison to $\text{MCA}_t=59.9$ on AWA2 of Zhang \etal \cite{zhang2017learning}, that projects word embeddings into the visual space to solve the hubness problem and uses a non-linear embedding function with a $L2$ loss, \csnl{} achieves $\text{MCA}_t=61.0$ and $\text{MCA}_t=71.4$-- in the transductive setting. \csnl{} also outperforms LATEM \cite{xian2016latent}, that uses a mixture of linear projections to learn a non-linear embedding space on AWA2, CUB, and Kinetics-ZS (e.g. $\text{MCA}_t=19.7$ versus $25.1$ on CUB and $51.0$ versus $67.5$ on Kinetics-ZS). This shows that \csnl{} can improve performance despite using a simpler (linear) embedding space and possibility yield further improvements when combined with more sophisticated projections. We also implement a variant of LATEM (LATEM-Hier) that combines word embeddings with hierarchical embeddings, derived from WordNet, for class embeddings \cite{xian2016latent}. 
\csnl{} reports better ($\text{MCA}_t=61.0$ versus $57.7$ on AWA2) numbers by effectively encoding knowledge from both hypernyms and attributes. 


We also observe consistent improvements on the GZSL. Most of the models achieve zero performance on the $\textbf{HM}$ metric since the VSE based methods are biased towards the seen classes. 
Only Zhang \cite{zhang2017learning}, Learning to Compare \cite{sung2018learning}, and QFSL$^{\text{tr}}$ \cite{song2018transductive} achieve non-trivial performance on the generalized task (\eg{} $\text{HM}=35.7$, $28.1$, and $42.2$ on AWA2 respectively). In comparison \csnl$^{\text{tr}}$ (variant using unlabeled test data) achieves a significantly higher performance with $\text{HM}=64.0$. We also observe similar trends on other datasets where \csnl$^{\text{tr}}$ achieves $\text{HM}=29.4$ (CUB) and $\text{HM}=47.0$ (Kinetics-ZS) versus $19.9$ and $41.9$ of QFSL$^{\text{tr}}$, when both models operate in the transductive setting. This supports our earlier results that \csnl{} can additionally regularize the embedding space by enforcing rules on unlabeled test data. Based on these gains, we conclude that \csnl{} improves the discriminability of the embedding space by enforcing logical rules expressing common-sense knowledge.

\section{Conclusion}

We improved zero shot learning (ZSL) by injecting common-sense rules in visual-semantic embedding (VSE). We achieved this through    
a novel neuro-symbolic loss that enforces logical rules expressing common-sense relationships over visual features. 
For improved learning with logical rules we proposed two key novelties over prior neuro-symbolic methods-- enforcing rules for a group instead of a single concept and confidence margins that enable implicit curriculum learning. 
Through detailed ablation studies we demonstrated
the benefits of injecting hypernym and attribute based rules in VSE on three ZSL datasets. We also observed that our loss can exploit unlabeled test data (in the transductive setting) to further improve performance.  
We finally showed consistent improvements over SOTA methods on both conventional and generalized ZSL. We plan to extend our work to other settings such as few-shot learning in the future.  

\section{Acknowledgement}
The authors would like to acknowledge Karen Myers, Bill Mark, and Rodrigo Braz for helpful discussions.

{\small
\bibliographystyle{ieee_fullname}
\bibliography{egbib}

\begin{thebibliography}{10}\itemsep=-1pt

\bibitem{achlioptas2001lower}
Dimitris Achlioptas.
\newblock Lower bounds for random 3-sat via differential equations.
\newblock {\em Theoretical Computer Science}, 265(1-2):159--185, 2001.

\bibitem{akata2015label}
Zeynep Akata, Florent Perronnin, Zaid Harchaoui, and Cordelia Schmid.
\newblock Label-embedding for image classification.
\newblock {\em IEEE transactions on pattern analysis and machine intelligence},
  38(7):1425--1438, 2015.

\bibitem{akata2015evaluation}
Zeynep Akata, Scott Reed, Daniel Walter, Honglak Lee, and Bernt Schiele.
\newblock Evaluation of output embeddings for fine-grained image
  classification.
\newblock In {\em Conference on computer vision and pattern recognition}, pages
  2927--2936, 2015.

\bibitem{al2016recovering}
Ziad Al-Halah, Makarand Tapaswi, and Rainer Stiefelhagen.
\newblock Recovering the missing link: Predicting class-attribute associations
  for unsupervised zero-shot learning.
\newblock In {\em Conference on computer vision and pattern recognition}, pages
  5975--5984, 2016.

\bibitem{bansal2018zero}
Ankan Bansal, Karan Sikka, Gaurav Sharma, Rama Chellappa, and Ajay Divakaran.
\newblock Zero-shot object detection.
\newblock In {\em European conference on computer vision}, pages 384--400,
  2018.

\bibitem{barz2019hierarchy}
Bj{\"o}rn Barz and Joachim Denzler.
\newblock Hierarchy-based image embeddings for semantic image retrieval.
\newblock In {\em Winter conference on applications of computer vision}, pages
  638--647, 2019.

\bibitem{benaim2018one}
Sagie Benaim and Lior Wolf.
\newblock One-shot unsupervised cross domain translation.
\newblock In {\em Neural information processing systems}, pages 2104--2114,
  2018.

\bibitem{bengio2009curriculum}
Yoshua Bengio, J{\'e}r{\^o}me Louradour, Ronan Collobert, and Jason Weston.
\newblock Curriculum learning.
\newblock In {\em International conference on machine learning}, pages 41--48,
  2009.

\bibitem{carreira2018short}
Joao Carreira, Eric Noland, Andras Banki-Horvath, Chloe Hillier, and Andrew
  Zisserman.
\newblock A short note about kinetics-600.
\newblock {\em arXiv preprint arXiv:1808.01340}, 2018.

\bibitem{chao1986probabilistic}
Ming-Te Chao and John Franco.
\newblock Probabilistic analysis of two heuristics for the 3-satisfiability
  problem.
\newblock {\em SIAM Journal on Computing}, 15(4):1106--1118, 1986.

\bibitem{chao2016empirical}
Wei-Lun Chao, Soravit Changpinyo, Boqing Gong, and Fei Sha.
\newblock An empirical study and analysis of generalized zero-shot learning for
  object recognition in the wild.
\newblock In {\em European conference on computer vision}, pages 52--68, 2016.

\bibitem{chen2016xgboost}
Tianqi Chen and Carlos Guestrin.
\newblock Xgboost: A scalable tree boosting system.
\newblock In {\em International conference on knowledge discovery and data
  mining}, pages 785--794, 2016.

\bibitem{dara2002clustering}
Rozita Dara, Stefan~C Kremer, and Deborah~A Stacey.
\newblock Clustering unlabeled data with soms improves classification of
  labeled real-world data.
\newblock In {\em International joint conference on neural networks}, volume~3,
  pages 2237--2242, 2002.

\bibitem{deng2009imagenet}
Jia Deng, Wei Dong, Richard Socher, Li-Jia Li, Kai Li, and Li Fei-Fei.
\newblock Imagenet: A large-scale hierarchical image database.
\newblock In {\em Conference on computer vision and pattern recognition}, pages
  248--255, 2009.

\bibitem{diligenti2017integrating}
Michelangelo Diligenti, Soumali Roychowdhury, and Marco Gori.
\newblock Integrating prior knowledge into deep learning.
\newblock In {\em International conference on machine learning and
  applications}, pages 920--923, 2017.

\bibitem{donadello2017logic}
Ivan Donadello, Luciano Serafini, and Artur~D'Avila Garcez.
\newblock Logic tensor networks for semantic image interpretation.
\newblock {\em arXiv preprint arXiv:1705.08968}, 2017.

\bibitem{feichtenhofer2018slowfast}
Christoph Feichtenhofer, Haoqi Fan, Jitendra Malik, and Kaiming He.
\newblock Slowfast networks for video recognition.
\newblock {\em arXiv preprint arXiv:1812.03982}, 2018.

\bibitem{frome2013devise}
Andrea Frome, Greg~S Corrado, Jon Shlens, Samy Bengio, Jeff Dean, Marc'Aurelio
  Ranzato, and Tomas Mikolov.
\newblock Devise: A deep visual-semantic embedding model.
\newblock In {\em Neural information processing systems}, pages 2121--2129,
  2013.

\bibitem{fung2003knowledge}
Glenn~M Fung, Olvi~L Mangasarian, and Jude~W Shavlik.
\newblock Knowledge-based support vector machine classifiers.
\newblock In {\em Neural information processing systems}, pages 537--544, 2003.

\bibitem{goldberg2009multi}
Andrew Goldberg, Xiaojin Zhu, Aarti Singh, Zhiting Xu, and Robert Nowak.
\newblock Multi-manifold semi-supervised learning.
\newblock In {\em Artificial intelligence and statistics}, pages 169--176,
  2009.

\bibitem{hariharan2017low}
Bharath Hariharan and Ross Girshick.
\newblock Low-shot visual recognition by shrinking and hallucinating features.
\newblock In {\em Conference on computer vision and pattern recognition}, pages
  3018--3027, 2017.

\bibitem{hu2018few}
Zikun Hu, Xiang Li, Cunchao Tu, Zhiyuan Liu, and Maosong Sun.
\newblock Few-shot charge prediction with discriminative legal attributes.
\newblock In {\em International conference on computational linguistics}, pages
  487--498, 2018.

\bibitem{hu2016harnessing}
Zhiting Hu, Xuezhe Ma, Zhengzhong Liu, Eduard Hovy, and Eric Xing.
\newblock Harnessing deep neural networks with logic rules.
\newblock {\em arXiv preprint arXiv:1603.06318}, 2016.

\bibitem{hu2016deep}
Zhiting Hu, Zichao Yang, Ruslan Salakhutdinov, and Eric Xing.
\newblock Deep neural networks with massive learned knowledge.
\newblock In {\em Conference on empirical methods in natural language
  processing}, pages 1670--1679, 2016.

\bibitem{huang2019generative}
He Huang, Changhu Wang, Philip~S Yu, and Chang-Dong Wang.
\newblock Generative dual adversarial network for generalized zero-shot
  learning.
\newblock In {\em Conference on computer vision and pattern recognition}, pages
  801--810, 2019.

\bibitem{jayaraman2014zero}
Dinesh Jayaraman and Kristen Grauman.
\newblock Zero-shot recognition with unreliable attributes.
\newblock In {\em Neural information processing systems}, pages 3464--3472,
  2014.

\bibitem{jiang2018hybrid}
Chenhan Jiang, Hang Xu, Xiaodan Liang, and Liang Lin.
\newblock Hybrid knowledge routed modules for large-scale object detection.
\newblock In {\em Neural information processing systems}, pages 1552--1563,
  2018.

\bibitem{joulin2016fasttext}
Armand Joulin, Edouard Grave, Piotr Bojanowski, Matthijs Douze, H{\'e}rve
  J{\'e}gou, and Tomas Mikolov.
\newblock Fasttext.zip: Compressing text classification models.
\newblock {\em arXiv preprint arXiv:1612.03651}, 2016.

\bibitem{kankuekul2012online}
Pichai Kankuekul, Aram Kawewong, Sirinart Tangruamsub, and Osamu Hasegawa.
\newblock Online incremental attribute-based zero-shot learning.
\newblock In {\em Conference on computer vision and pattern recognition}, pages
  3657--3664, 2012.

\bibitem{kay2017kinetics}
Will Kay, Joao Carreira, Karen Simonyan, Brian Zhang, Chloe Hillier, Sudheendra
  Vijayanarasimhan, Fabio Viola, Tim Green, Trevor Back, Paul Natsev, Mustafa
  Suleyman, and Andrew Zisserman.
\newblock The kinetics human action video dataset.
\newblock {\em arXiv preprint arXiv:1705.06950}, 2017.

\bibitem{kingma2014semi}
Durk~P Kingma, Shakir Mohamed, Danilo~Jimenez Rezende, and Max Welling.
\newblock Semi-supervised learning with deep generative models.
\newblock In {\em Neural information processing systems}, pages 3581--3589,
  2014.

\bibitem{kodirov2015unsupervised}
Elyor Kodirov, Tao Xiang, Zhenyong Fu, and Shaogang Gong.
\newblock Unsupervised domain adaptation for zero-shot learning.
\newblock In {\em International conference on computer vision}, pages
  2452--2460, 2015.

\bibitem{lampert2013attribute}
Christoph~H Lampert, Hannes Nickisch, and Stefan Harmeling.
\newblock Attribute-based classification for zero-shot visual object
  categorization.
\newblock {\em IEEE transactions on pattern analysis and machine intelligence},
  36(3):453--465, 2013.

\bibitem{le2014distributed}
Quoc Le and Tomas Mikolov.
\newblock Distributed representations of sentences and documents.
\newblock In {\em International conference on machine learning}, pages
  1188--1196, 2014.

\bibitem{li2017learning}
Dong Li, Hsin-Ying Lee, Jia-Bin Huang, Shengjin Wang, and Ming-Hsuan Yang.
\newblock Learning structured semantic embeddings for visual recognition.
\newblock {\em arXiv preprint arXiv:1706.01237}, 2017.

\bibitem{li2019leveraging}
Jingjing Li, Mengmeng Jing, Ke Lu, Zhengming Ding, Lei Zhu, and Zi Huang.
\newblock Leveraging the invariant side of generative zero-shot learning.
\newblock In {\em Conference on computer vision and pattern recognition}, pages
  7402--7411, 2019.

\bibitem{liang2018symbolic}
Xiaodan Liang, Zhiting Hu, Hao Zhang, Liang Lin, and Eric~P Xing.
\newblock Symbolic graph reasoning meets convolutions.
\newblock In {\em ANeural information processing systems}, pages 1853--1863,
  2018.

\bibitem{lin2014microsoft}
Tsung-Yi Lin, Michael Maire, Serge Belongie, James Hays, Pietro Perona, Deva
  Ramanan, Piotr Doll{\'a}r, and C~Lawrence Zitnick.
\newblock Microsoft coco: Common objects in context.
\newblock In {\em European conference on computer vision}, pages 740--755,
  2014.

\bibitem{luo2017label}
Zelun Luo, Yuliang Zou, Judy Hoffman, and Li~F Fei-Fei.
\newblock Label efficient learning of transferable representations acrosss
  domains and tasks.
\newblock In {\em Neural information processing systems}, pages 165--177, 2017.

\bibitem{mangasarian2007nonlinear}
Olvi~L Mangasarian and Edward~W Wild.
\newblock Nonlinear knowledge in kernel approximation.
\newblock {\em IEEE transactions on neural networks}, 18(1):300--306, 2007.

\bibitem{miller1998wordnet}
George~A Miller.
\newblock {\em WordNet: An electronic lexical database}.
\newblock MIT press, 1998.

\bibitem{morgado2017semantically}
Pedro Morgado and Nuno Vasconcelos.
\newblock Semantically consistent regularization for zero-shot recognition.
\newblock In {\em Conference on computer vision and pattern recognition}, pages
  6060--6069, 2017.

\bibitem{motiian2017few}
Saeid Motiian, Quinn Jones, Seyed Iranmanesh, and Gianfranco Doretto.
\newblock Few-shot adversarial domain adaptation.
\newblock In {\em Neural information processing systems}, pages 6670--6680,
  2017.

\bibitem{norouzi2013zero}
Mohammad Norouzi, Tomas Mikolov, Samy Bengio, Yoram Singer, Jonathon Shlens,
  Andrea Frome, Greg~S Corrado, and Jeffrey Dean.
\newblock Zero-shot learning by convex combination of semantic embeddings.
\newblock {\em arXiv preprint arXiv:1312.5650}, 2013.

\bibitem{odena2016semi}
Augustus Odena.
\newblock Semi-supervised learning with generative adversarial networks.
\newblock {\em arXiv preprint arXiv:1606.01583}, 2016.

\bibitem{paszke2019pytorch}
Adam Paszke, Sam Gross, Francisco Massa, Adam Lerer, James Bradbury, Gregory
  Chanan, Trevor Killeen, Zeming Lin, Natalia Gimelshein, Luca Antiga, et~al.
\newblock Pytorch: An imperative style, high-performance deep learning library.
\newblock In {\em Neural information processing systems}, pages 8026--8037,
  2019.

\bibitem{qiao2017visually}
Ruizhi Qiao, Lingqiao Liu, Chunhua Shen, and Anton van~den Hengel.
\newblock Visually aligned word embeddings for improving zero-shot learning.
\newblock {\em arXiv preprint arXiv:1707.05427}, 2017.

\bibitem{salimans2016improved}
Tim Salimans, Ian Goodfellow, Wojciech Zaremba, Vicki Cheung, Alec Radford, and
  Xi Chen.
\newblock Improved techniques for training gans.
\newblock In {\em Neural information processing systems}, pages 2234--2242,
  2016.

\bibitem{sariyildiz2019gradient}
Mert~Bulent Sariyildiz and Ramazan~Gokberk Cinbis.
\newblock Gradient matching generative networks for zero-shot learning.
\newblock In {\em Conference on computer vision and pattern recognition}, pages
  2168--2178, 2019.

\bibitem{scheirer2012toward}
Walter~J Scheirer, Anderson de Rezende~Rocha, Archana Sapkota, and Terrance~E
  Boult.
\newblock Toward open set recognition.
\newblock {\em IEEE transactions on pattern analysis and machine intelligence},
  35(7):1757--1772, 2012.

\bibitem{serafini2016logic}
Luciano Serafini and Artur~d'Avila Garcez.
\newblock Logic tensor networks: Deep learning and logical reasoning from data
  and knowledge.
\newblock {\em arXiv preprint arXiv:1606.04422}, 2016.

\bibitem{sheikhpour2017survey}
Razieh Sheikhpour, Mehdi~Agha Sarram, Sajjad Gharaghani, and Mohammad Ali~Zare
  Chahooki.
\newblock A survey on semi-supervised feature selection methods.
\newblock {\em Pattern recognition}, 64:141--158, 2017.

\bibitem{shyam2017attentive}
Pranav Shyam, Shubham Gupta, and Ambedkar Dukkipati.
\newblock Attentive recurrent comparators.
\newblock {\em arXiv preprint arXiv:1703.00767}, 2017.

\bibitem{sikka2020deep}
Karan Sikka, Andrew Silberfarb, John Byrnes, Indranil Sur, Ed Chow, Ajay
  Divakaran, and Richard Rohwer.
\newblock Deep adaptive semantic logic (dasl): Compiling declarative knowledge
  into deep neural networks.
\newblock {\em arXiv preprint arXiv:2003.07344}, 2020.

\bibitem{snell2017prototypical}
Jake Snell, Kevin Swersky, and Richard Zemel.
\newblock Prototypical networks for few-shot learning.
\newblock In {\em Advances in neural information processing systems}, pages
  4077--4087, 2017.

\bibitem{song2018transductive}
Jie Song, Chengchao Shen, Yezhou Yang, Yang Liu, and Mingli Song.
\newblock Transductive unbiased embedding for zero-shot learning.
\newblock In {\em Conference on computer vision and pattern recognition}, pages
  1024--1033, 2018.

\bibitem{stewart2017label}
Russell Stewart and Stefano Ermon.
\newblock Label-free supervision of neural networks with physics and domain
  knowledge.
\newblock In {\em AAAI conference on artificial intelligence}, 2017.

\bibitem{sun2017revisiting}
Chen Sun, Abhinav Shrivastava, Saurabh Singh, and Abhinav Gupta.
\newblock Revisiting unreasonable effectiveness of data in deep learning era.
\newblock In {\em International conference on computer vision}, pages 843--852,
  2017.

\bibitem{sung2018learning}
Flood Sung, Yongxin Yang, Li Zhang, Tao Xiang, Philip~HS Torr, and Timothy~M
  Hospedales.
\newblock Learning to compare: Relation network for few-shot learning.
\newblock In {\em Conference on computer vision and pattern recognition}, pages
  1199--1208, 2018.

\bibitem{triguero2015self}
Isaac Triguero, Salvador Garc{\'\i}a, and Francisco Herrera.
\newblock Self-labeled techniques for semi-supervised learning: taxonomy,
  software and empirical study.
\newblock {\em Knowledge and Information systems}, 42(2):245--284, 2015.

\bibitem{van2018inaturalist}
Grant Van~Horn, Oisin Mac~Aodha, Yang Song, Yin Cui, Chen Sun, Alex Shepard,
  Hartwig Adam, Pietro Perona, and Serge Belongie.
\newblock The inaturalist species classification and detection dataset.
\newblock In {\em Conference on computer vision and pattern recognition}, pages
  8769--8778, 2018.

\bibitem{verma2012learning}
Nakul Verma, Dhruv Mahajan, Sundararajan Sellamanickam, and Vinod Nair.
\newblock Learning hierarchical similarity metrics.
\newblock In {\em Conference on computer vision and pattern recognition}, pages
  2280--2287, 2012.

\bibitem{vinyals2016matching}
Oriol Vinyals, Charles Blundell, Timothy Lillicrap, Daan Wierstra, et~al.
\newblock Matching networks for one shot learning.
\newblock In {\em Neural information processing systems}, pages 3630--3638,
  2016.

\bibitem{wah2011caltech}
Catherine Wah, Steve Branson, Peter Welinder, Pietro Perona, and Serge
  Belongie.
\newblock The caltech-ucsd birds-200-2011 dataset.
\newblock 2011.

\bibitem{wang2019unifying}
Qian Wang, Penghui Bu, and Toby~P Breckon.
\newblock Unifying unsupervised domain adaptation and zero-shot visual
  recognition.
\newblock In {\em International joint conference on neural networks}, pages
  1--8, 2019.

\bibitem{wang2017learning}
Yu-Xiong Wang, Deva Ramanan, and Martial Hebert.
\newblock Learning to model the tail.
\newblock In {\em Neural information processing systems}, pages 7029--7039,
  2017.

\bibitem{xian2016latent}
Yongqin Xian, Zeynep Akata, Gaurav Sharma, Quynh Nguyen, Matthias Hein, and
  Bernt Schiele.
\newblock Latent embeddings for zero-shot classification.
\newblock In {\em Conference on computer vision and pattern recognition}, pages
  69--77, 2016.

\bibitem{xian2018feature}
Yongqin Xian, Tobias Lorenz, Bernt Schiele, and Zeynep Akata.
\newblock Feature generating networks for zero-shot learning.
\newblock In {\em Conference on computer vision and pattern recognition}, pages
  5542--5551, 2018.

\bibitem{xian2017zero}
Yongqin Xian, Bernt Schiele, and Zeynep Akata.
\newblock Zero-shot learning-the good, the bad and the ugly.
\newblock In {\em Conference on computer vision and pattern recognition}, pages
  4582--4591, 2017.

\bibitem{xu2018semantic}
Jingyi Xu, Zilu Zhang, Tal Friedman, Yitao Liang, and Guy Broeck.
\newblock A semantic loss function for deep learning with symbolic knowledge.
\newblock In {\em International conference on machine learning}, pages
  5502--5511, 2018.

\bibitem{yan2015multi}
Wang Yan, Jordan Yap, and Greg Mori.
\newblock Multi-task transfer methods to improve one-shot learning for
  multimedia event detection.
\newblock In {\em British Machine Vision Conference}, pages 37--1, 2015.

\bibitem{zhang2017learning}
Li Zhang, Tao Xiang, and Shaogang Gong.
\newblock Learning a deep embedding model for zero-shot learning.
\newblock In {\em Conference on computer vision and pattern recognition}, pages
  2021--2030, 2017.

\bibitem{zhang2018fine}
Yabin Zhang, Hui Tang, and Kui Jia.
\newblock Fine-grained visual categorization using meta-learning optimization
  with sample selection of auxiliary data.
\newblock In {\em European conference on computer vision}, pages 233--248,
  2018.

\bibitem{zhou2010semi}
Zhi-Hua Zhou and Ming Li.
\newblock Semi-supervised learning by disagreement.
\newblock {\em Knowledge and information systems}, 24(3):415--439, 2010.

\bibitem{zhu2018generative}
Yizhe Zhu, Mohamed Elhoseiny, Bingchen Liu, Xi Peng, and Ahmed Elgammal.
\newblock A generative adversarial approach for zero-shot learning from noisy
  texts.
\newblock In {\em Conference on computer vision and pattern recognition}, pages
  1004--1013, 2018.

\bibitem{zhu2019learning}
Yizhe Zhu, Jianwen Xie, Bingchen Liu, and Ahmed Elgammal.
\newblock Learning feature-to-feature translator by alternating
  back-propagation for generative zero-shot learning.
\newblock In {\em Conference on computer vision and pattern recognition}, pages
  9844--9854, 2019.

\end{thebibliography}
}

\newpage
\section{Handling Local Minima in \csnl{} Loss}

One can address the creation of unwanted local minima induced by the logical rules in several ways. For example, it can achieved by using a scheduler to slowly \colons{turn on} the logical loss, by using a full probabilistic formulation, or by using a curriculum to first train using the data and then begin enforcing rules in some priority order. All of these approaches have noticeable drawbacks. Scheduling of the weights of the rules is problematic because the implication based rules can impact the correct application of subsequent rules (e.g. rule 1 says that the animal is \colons{White} then rule 2 relies on that information to learn to predict \colons{Polar Bear} rather than \colons{Brown Bear}). Using a curriculum for enforcing rules can solve this issue, but requires someone to explicitly develop the curriculum, and removes the possibility of using rule application on simple cases to inform how rules should be applied on more challenging examples. Finally, a fully probabilistic approach works well in relatively simple cases by effectively solving the constrains ahead of time, but fails when constraint satisfaction intimately depends on the data. Our approach of setting confidence margins on the rules closely impersonates curriculum learning (referred to as implicit curriculum in the main text) strategy of choosing simpler to difficult samples for learning based on the confidence of each data sample satisfies the rules. This avoids us from the developing multi-stage data strategies that is required in standard curriculum learning applications. 

\section{Choice of Hierarchy}
\begin{figure*}[tb]
    \centering{\includegraphics[width=1\columnwidth]{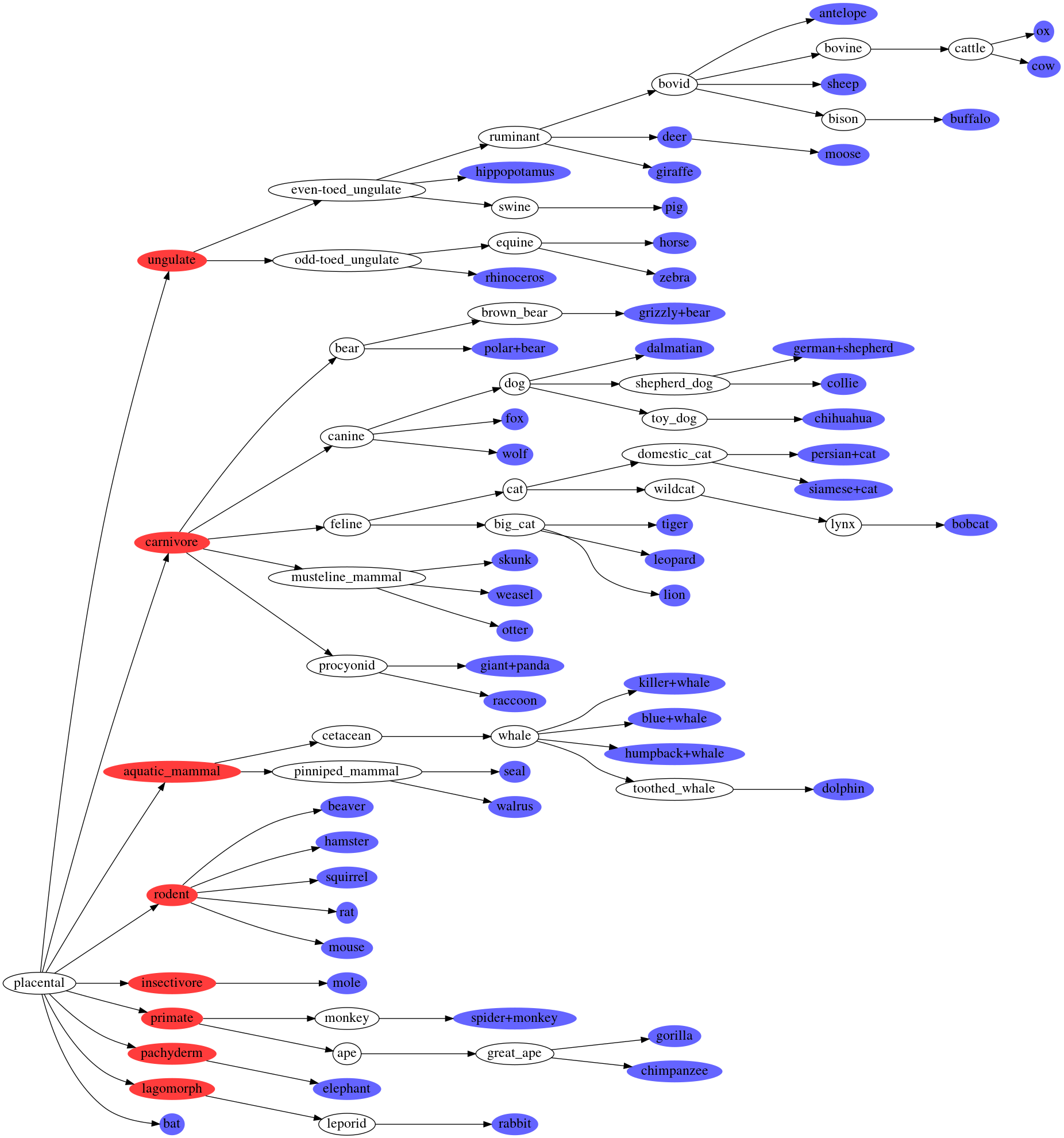}}

    \centering{\includegraphics[width=1\columnwidth]{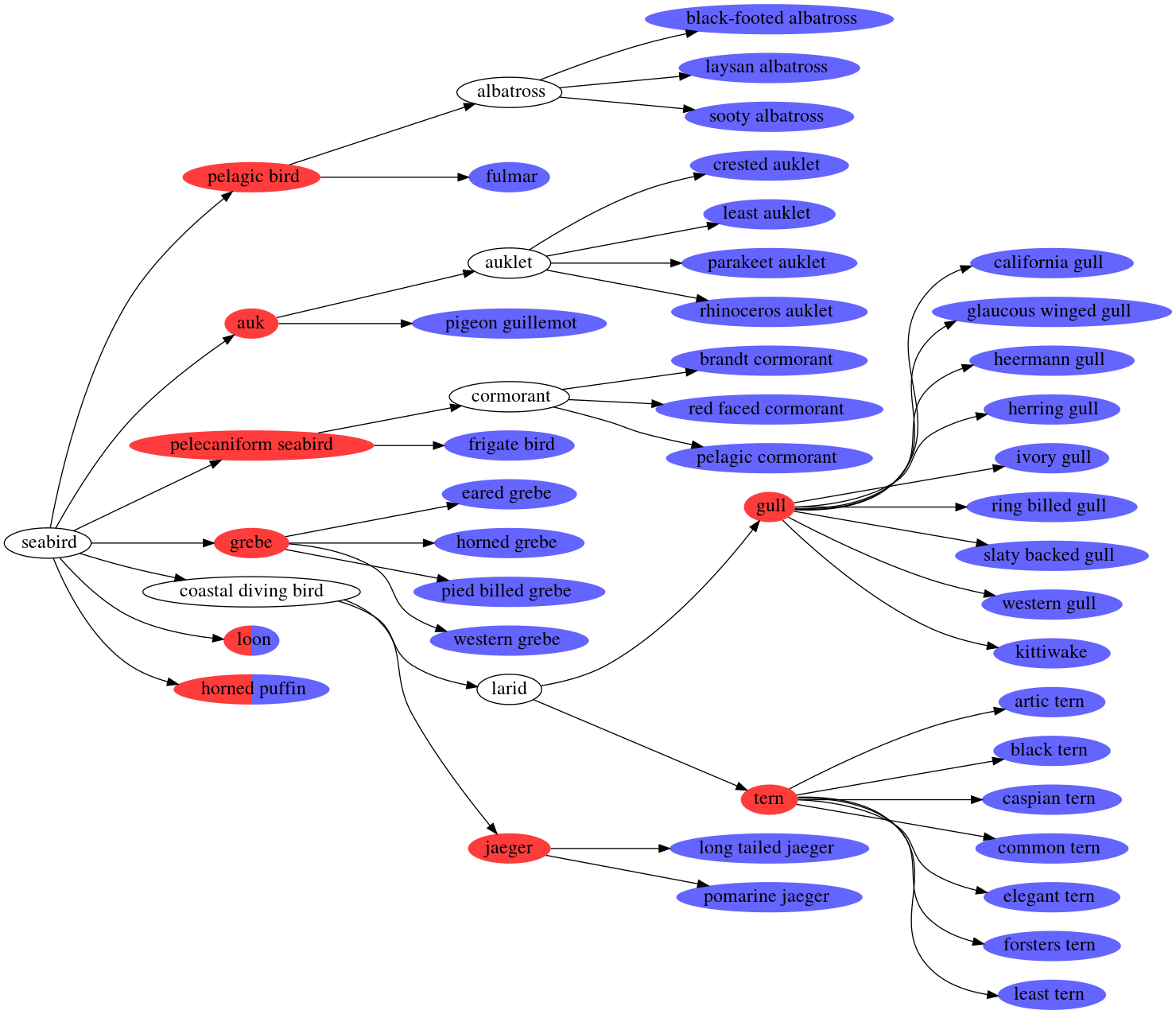}}
    \caption{The knowledge hierarchy derived from WordNet for AWA2 and CUB as used in \csnl{}. 
             Top: AWA2. Bottom: CUB.
             Blue nodes are classes in the dataset, and red nodes are selected hypernyms. Only a part of the entire hierarchy is shown due to space limitation.
             Best viewed in color.}
    \label{fig:class_hierarchy}
\end{figure*}

\autoref{fig:class_hierarchy} shows the knowledge hierarchy derived from WordNet for AWA2 and CUB as used in \csnl{} (see implementation details in \autoref{sec:experiments} in the main text). We used top-level synsets \colons{Placental} and \colons{Seabird} for AWA2 and CUB respectively as these had decent coverage for the all the dataset classes. We also experimented with additional top-level synsets such a \colons{Feline}, but the improvement in performance was minimal. 

\section{Parametric Study}
\subsection{Hyperparameters for Logical Loss}

\begin{figure*}[htbp]
    \centering{\includegraphics[width=.23\textwidth]{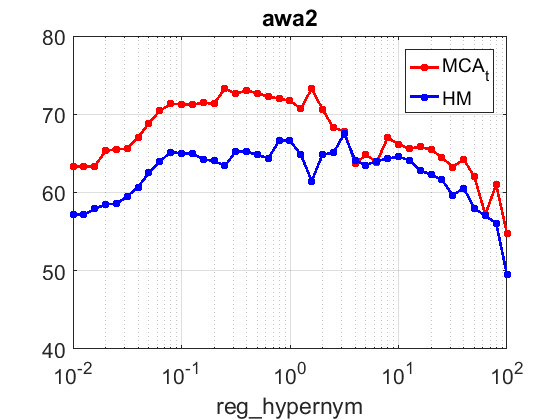}}\hfill
    \centering{\includegraphics[width=.23\textwidth]{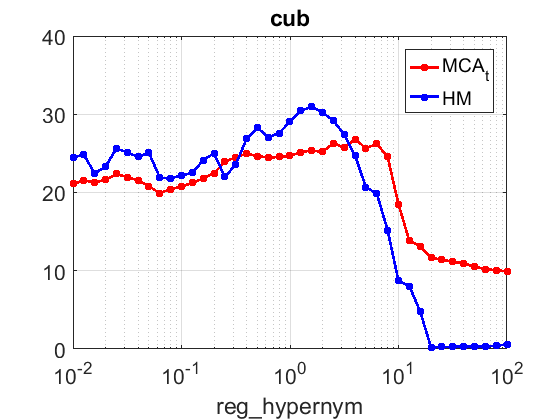}}\hfill
    \centering{\includegraphics[width=.23\textwidth]{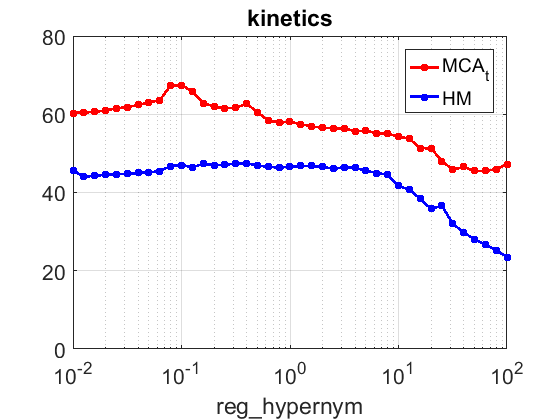}}\hfill    
    \caption{
    Effect of regularization parameter ($\lambda_{\text{hyp}}$) for the hypernym based logical loss on $\text{MCA}_t$ and $\text{HM}$ for all the datasets.
             In each subfigure X$-$axis is the regularization parameter in log scale, and Y$-$axis is the performance metric. 
             Best viewed in color.}
    \label{fig:hyper}
\end{figure*}

As discussed in \autoref{sec:cnsl} and shown below, \csnl{} uses regularization parameters for logical losses corresponding to hypernyms and attributes. 
\begin{align}
    \L &= \L_t + \L_{\text{\csnl}} (\mathcal{D}^s) + \lambda_{\text{trans}}\L_{\text{\csnl}} (\mathcal{D}^{\text{trans}})  \\
    \L_{\text{\csnl}}(\mathcal{D}) &= \sum_{i=1}^{N} (\lambda_{\text{hyp}} L_{\text{hyp}}(x_i) + \lambda_{\text{attr}} L_{\text{attr}}(x_i)) 
\end{align}

We set these hyperparameters using cross-validation. \autoref{fig:hyper} shows the effect of $\lambda_{\text{hyp}}$ on performance. 
As expected, the performance first improves on increasing $\lambda_{\text{hyp}}$ and then drops for higher values of $\lambda_{\text{hyp}}$. This is expected since the loss will not be active for very low values of $\lambda_{\text{hyp}}$ and will result in overfitting for very high values of $\lambda_{\text{hyp}}$. 


\subsection{Confidence Sweep}

\begin{figure*}[htbp]
    \centering{\includegraphics[width=.47\columnwidth]{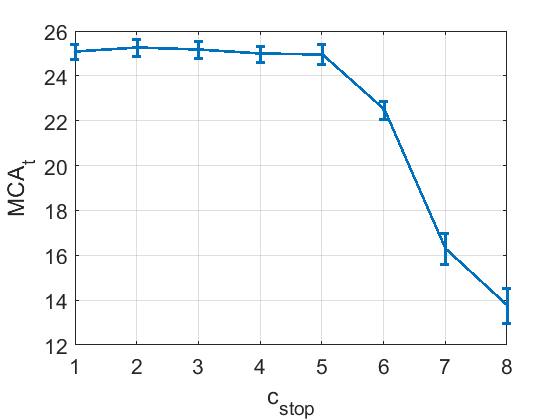}}
    \centering{\includegraphics[width=.47\columnwidth]{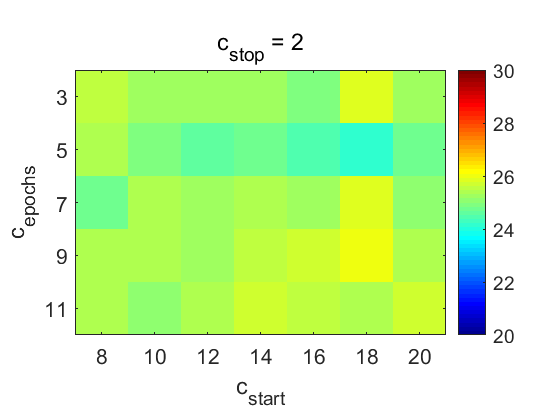}}
    \caption{Effect of hyperparameters controlling the confidence margin in \csnl{} on zero-shot classification on CUB. Left plot shows averaged performance for $c_\text{stop}$ while keeping $c_\text{start}$ and $c_\text{epochs}$ fixed. Right plot shows 2D plot for $c_\text{start}$ and $c_\text{epochs}$ while keeping $c_\text{stop}$ fixed (best seen in color).} 
    \label{fig:conf_cub}
\end{figure*}

\begin{figure*}[htbp]
    \centering{\includegraphics[width=.47\columnwidth]{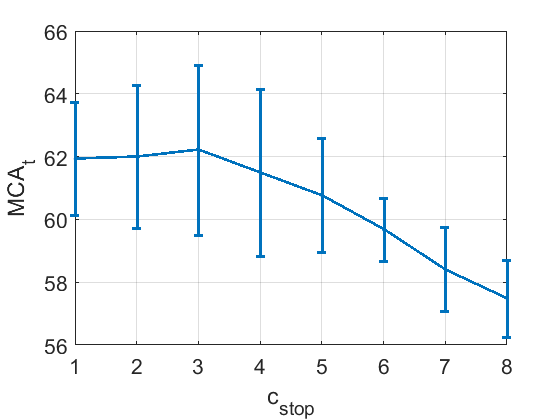}}
    \centering{\includegraphics[width=.47\columnwidth]{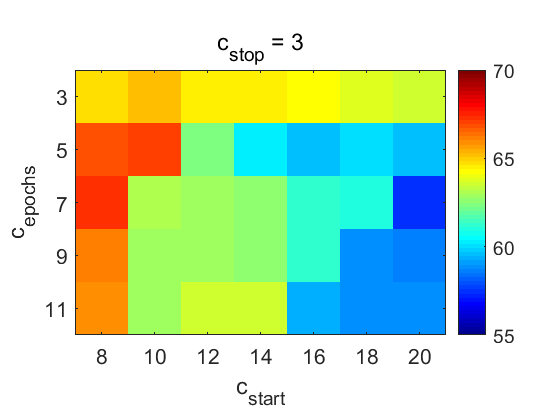}}
    \caption{Effect of hyperparameters controlling the confidence margin in \csnl{} on zero-shot classification on Kinetics-ZS. Left plot shows averaged performance for $c_\text{stop}$ while keeping $c_\text{start}$ and $c_\text{epochs}$ fixed. Right plot shows 2D plot for $c_\text{start}$ and $c_\text{epochs}$ while keeping $c_\text{stop}$ fixed (best seen in color).} 
    \label{fig:conf_kinetics}
\end{figure*}

Similar to \autoref{sec:conf_margin}, \autoref{fig:conf_cub} and \autoref{fig:conf_kinetics} show the effect of confidence margin parameters on the $\text{MCA}_t$ for CUB and Kinetics-ZS respectively. 
The trends are similar to AWA2. For example, the performance drops for higher values of $c_\text{stop}$ as the rules are not enforced in this case.  
In the case of CUB, the lower $c_\text{stop}$ values don't really show a sign of overfitting as seen in AWA2 and this could be due to better coherence in the confidence of early predictions and satisfaction on the rules. For Kinetics-ZS the performance is high for $c_\text{start}=8$ and $5<=c_\text{epochs}<=7$ or $c_\text{start}=9$ and $c_\text{epochs}=5$. There are similarities with the trend for AWA2, where we observed an inverse relationship between $c_\text{start}$ and $c_\text{epochs}$ for achieving high performance. However, we advise to set these parameters using cross-validation due to dataset specific biases.

\section{Implementation details for SOTA models}
\label{sec:implementation}
To evaluate SOTA methods, we use the same features and train/test splits as in our method. We 
implement SJE \cite{akata2015evaluation}, Latem \cite{xian2016latent}, Latem-Hier \cite{xian2016latent}, Devise \cite{frome2013devise} and Devise (image space) \cite{frome2013devise} using triplet loss as described in the papers.
For Latem-Hier \cite{xian2016latent} we use the hierarchical features for CUB and AWA2 provided by its author. We did not implement Latem-Hier for Kinetics-ZS since the hierarchical features were not available.  
We use the source code of Zhang \cite{zhang2017learning} and Learning to Compare \cite{sung2018learning} provided by their authors.
We also implement the transductive loss in QFSL$^{\text{tr}}$ \cite{song2018transductive} as our baseline in the transductive setting.
For each SOTA method we search for its hyperparameters around the values mentioned in its paper and pick the ones that give the best performance on the test data.

\section{Details on Kinetics-ZS}
\label{sec:kinetics}
The \textbf{Kinetics-ZS}, derived from Kinetics 400 \cite{kay2017kinetics}, contains $91$ seen and $18$ unseen classes, which have been organized into $10$ mutually exclusive hypernyms. These classes and hypernyms are shown in \autoref{table:kinetics_classes}. We manually labeled $20$ attributes for Kinetics, which are shown in \autoref{table:kinetics_attributes}. 

Each YouTube video in the dataset is $10$ seconds long and has the same class label for the entirety of the video. To extract video features, we first split each $10$s video into five distinct $2$s clips before feeding each clip into the SlowFast network. Each clip is then resized and center-cropped to a $224 \times 224$ pixel square and flipped horizontally with $50\%$ probability. At the end of this pre-processing step, every unique video corresponds to between $5-7$ unique video clips (as the horizontal flipping allows for duplicate temporal overlaps). The SlowFast network uses a ResNet-50 backbone and extracts features for each of these 2s video clips at a temporal stride of $\tau =16$, a speed ratio of $\alpha = 8$, and a channel ratio of $\beta = \frac{1}{8}$. Hence, the slow pathway will process $T$ frames over $C$ channels, and the fast pathway will process $\alpha T$ frames over $\beta C$ channels, where $T = \frac{\text{frames per second} \times \text{2s}}{\tau}$. 

Our final dataset contains a training set, a testing set of seen classes, and a testing set of unseen classes. The training set consists of features from 11,311 video clips derived for 2,136 unique videos, the testing set of seen classes consists of features for 2,948 video clips from 1,725 unique videos, and the testing set of unseen classes consists of features for 2,346 video clips from 455 unique videos (70-15-15 split). We will release these splits as well as the hypernym and attribute information. 

\begin{table*}[htbp]
    \centering
    \begin{tabular}{c}
        \hline
        \textbf{Attributes} \\
        \hline
         has ball \\
         has racquet \\
         has bat \\
         has club or stick \\
         has helmet \\
         has bull \\
         has horse \\
         has animals \\
         has wheels \\
         has person riding \\
         has projectile \\
         has acrobatics \\
         has target \\
         holding or wearing no equipment \\
         in air falling or swinging \\
         in air from ground \\
         is on water \\
         is on snow or ice \\
         is on court \\
         is on field \\
    \end{tabular}
    \caption{Manually labeled attributes for the Kinetics-ZS dataset.}
    \label{table:kinetics_attributes}
\end{table*}

\begin{table*}[htbp!]
    \centering
    \begin{tabular}{cccc}
        \hline
        \textbf{Hypernym} & \textbf{Seen} & \textbf{Unseen} & \textbf{Classes} \\ 
        \hline
	\multirow{1}{*}{Athletics-Jumping} & \multirow{1}{*}{6} & \multirow{1}{*}{0}
		&  high jump, hurdling, long jump, parkour, pole vault, triple jump \\ 
	\hline
        \multirow{3}{*}{Athletics-Throwing+Launching} & \multirow{3}{*}{9} & \multirow{3}{*}{1}
        		&  archery, catching or throwing frisbee, disc golfing, hammer throw, \\
		& & &  javelin throw, shot put, throwing axe, throwing ball, \\ 
		& & & throwing discus, \textit{playing darts} \\
	\hline
	\multirow{8}{*}{Ball Sports} & \multirow{8}{*}{17} & \multirow{8}{*}{2}
        		&  bowling, dodgeball, dribbling basketball, dunking basketball, \\
		& & &  golf chipping, golf driving, golf putting, juggling soccer ball, \\
		& & &  kicking field goal, kicking soccer ball, \\
		& & &  passing American football (in game), \\
		& & &  passing American football (not in game), \\
		& & &  playing basketball, playing kickball, playing volleyball, \\
		& & &  shooting basketball, shooting goal (soccer), \\ 
		& & & \textit{passing soccer ball}, \textit{playing netball} \\
	\hline
	\multirow{2}{*}{Gymnastics} & \multirow{2}{*}{5} & \multirow{2}{*}{0}
		&  bouncing on trampoline, cartwheeling, gymnastics tumbling, \\
		& & & somersaulting, vault \\
	\hline
	\multirow{3}{*}{Heights} & \multirow{3}{*}{10} & \multirow{3}{*}{1}
		&  abseiling, bungee jumping, diving cliff, ice climbing, paragliding, \\
		& & &  rock climbing, skydiving, slacklining, springboard diving, \\
		& & & trapezing, \textit{base jumping} \\
	\hline	
	\multirow{3}{*}{Mobility-Land} & \multirow{3}{*}{8} & \multirow{3}{*}{3}
		&  jogging, motorcycling, riding a bike, riding mountain bike, \\
		& & &  riding scooter, riding unicycle, roller skating, skateboarding, \\
		& & & \textit{jumping bicycle}, \textit{longboarding}, \textit{bullfighting} \\
	\hline	
	\multirow{3}{*}{Mobility-Water} & \multirow{3}{*}{6} & \multirow{3}{*}{2}
		&  scuba diving, snorkeling, swimming backstroke, \\
		& & &  swimming breaststroke, swimming butterfly stroke, \\
		& & & water sliding, \textit{ice swimming}, \textit{swimming front crawl} \\
	\hline
	\multirow{5}{*}{Racquet+Bat Sports} & \multirow{5}{*}{8} & \multirow{5}{*}{5}
		&  catching or throwing baseball, catching or throwing softball, \\
		& & &  hurling (sport), playing badminton, playing cricket, \\
		& & & playing squash or racquetball, playing tennis, \textit{fencing (sport)} \\
		& & & \textit{playing field hocket}, \textit{playing ping pong}, \\
		& & & \textit{playing polo}, \textit{swinging baseball bat} \\
	\hline
	\multirow{6}{*}{Snow+Ice} & \multirow{6}{*}{14} & \multirow{6}{*}{3}
		&  bobsledding, hocket stop, ice fishing, ice skating, \\
		& & &  playing ice hockey, ski jumping, \\
		& & & skiing (not slalom or crosscountry), skiing crosscountry,  \\
		& & & skiing slalom, sled dog racing, snowboarding, snowkiting, \\
		& & & snowmobiling, tobogganing, \textit{curling(sport)}, \\
		& & & \textit{luge}, \textit{skiing mono} \\
	\hline
	\multirow{2}{*}{Water Sports} & \multirow{2}{*}{8} & \multirow{2}{*}{1}
		&  canoeing or kayaking, jetskiing, kitesurfing, parasailing, sailing, \\
		& & &  surfing water, water skiing, windsurfing, \textit{bodysurfing} \\
	\hline
        
    \end{tabular}
    \caption{Class and Hypernym names for the Kinetics-ZS dataset. Unseen classes are shown in \textit{italics}.}
    \label{table:kinetics_classes}
\end{table*}

\end{document}